\documentclass{article}

\usepackage[final]{neurips_2024}

\usepackage{amsmath,amssymb,amsthm}
\usepackage{algorithm}
\usepackage{algpseudocode}
\usepackage{hyperref}
\usepackage{graphicx}
\usepackage{booktabs}
\usepackage{natbib}
\usepackage{url}
\usepackage{placeins}
\usepackage{multirow}
\usepackage{float}
\usepackage{subcaption}
\usepackage{listings}
\usepackage{threeparttable}
\usepackage[most]{tcolorbox}

\makeatletter
\renewcommand{\@notice}{}
\makeatother

\title{Variational Boosting for Physics-Informed Neural Networks}

\author{
Kaylee Vo \\
Harvard University \\
\texttt{kav418@g.harvard.edu}
\And
Pavlos Protopapas \\
Harvard University \\
\texttt{pavlos@seas.harvard.edu}
}

\begin{document}

\maketitle

\begin{abstract}
Physics-Informed Neural Networks (PINNs) solve differential equations by minimizing the residual of a nonlinear operator over a neural parameterization of the solution. However, monolithic PINNs often suffer from ill-conditioning, spectral bias, and optimization instability.

We introduce a variational boosting framework in which solutions are constructed additively in function space. Each stage trains a weak learner whose converged correction satisfies a local orthogonality condition, equivalent to a projected functional gradient descent step onto the tangent space of the network's function manifold. Because each correction network is deliberately small, the restricted minimization admits full Newton or conjugate gradient updates, which are typically infeasible in large PINNs. The resulting method separates global nonlinear refinement into a sequence of well-conditioned subproblems while preserving the full variational structure of the operator.

This framework provides a geometric interpretation of multi-stage PINNs as projected functional gradient descent and enables stable second-order optimization for nonlinear differential equations.
\end{abstract}

\section{Introduction}
\label{section:introduction}
Physics-Informed Neural Networks (PINNs) approximate solutions of differential equations by minimizing a residual functional
\begin{equation}
    \mathcal L(u) = \|F(u)\|^2,
\end{equation}
alongside boundary and initial conditions (\cite{Raissi2019PINNs}; \cite{karniadakis-2021}). $F$ is a nonlinear differential operator, and $\|\cdot\|^2$ is a discrete $L^2$ norm evaluated at a set of collocation points scattered across the domain. The classical formulation minimizes $\mathcal L(u_\theta)$ over a neural parameterization $u_\theta$: the solution $u$ is the function we seek, living in the infinite-dimensional space $H^m$, while $\theta \in \mathbb{R}^p$ is a finite-dimensional vector of neural network parameters (e.g., weights and biases $\{W_1, W_2, b_1, b_2\}$ of a network $u_\theta(x,t) = W_2\cdot\sigma(W_1\cdot[x,t] + b_1) + b_2$), and $u_\theta$ denotes the function represented by the network with parameters $\theta$. The term ``parameterization'' emphasizes that this finite-dimensional vector $\theta$ is used to search the infinite-dimensional function space in which $u$ resides.
Classical PINNs solve
\begin{equation}
\min_\theta \mathcal L(u_\theta),
\end{equation}
performing descent in the parameter space of a large neural network.
This optimization problem is highly nonlinear, as the mapping $\theta \mapsto u_\theta$ is nonlinear and typically overparameterized. Moreover, all scales and nonlinear interactions of the solution are entangled within a single global model, often leading to ill-conditioning (\cite{wang-2021}), spectral bias (~\cite{pmlr-v97-rahaman19a}), and unstable training dynamics.

\paragraph{Motivation} To better understand these challenges, it is useful to examine how the residual changes under perturbations of the function $u$ itself. The functional gradient of the residual satisfies
\begin{equation}
    \nabla_u \mathcal L(u) = 2 F'(u)^* F(u),
    \label{eq:functional_gradient_closed_form}
\end{equation}
where $F'(u)$ denotes the Fréchet derivative of $F$ at $u$. It is the linear operator that best approximates how $F$ changes under small perturbations of $u$. For a small perturbation $h$,
\begin{equation}
    F(u+h) \approx F(u) + F'(u)[h],
\end{equation}
where $F'(u)[h]$ denotes applying the linear operator $F'(u)$ to the function $h$; crucially, $F'(u)$ is linear in $h$ even though $F$ itself may be highly nonlinear in $u$. The notation $F'(u)$ denotes the derivative evaluated at the point $u$.

$F'(u)^*$ is the adjoint of the Fréchet derivative, which maps the residuals back into the same space as $u$. Concretely,
\begin{equation}
    F'(u) \colon \mathcal{H} \to L^2, \qquad F'(u)^* \colon L^2 \to \mathcal{H},
\end{equation}
so $F'(u)$ maps functions to residuals, while its adjoint $F'(u)^*$ maps residuals back into function space. The two are related by the defining inner-product identity
\begin{equation}
    \langle F'(u)[v], w \rangle_{L^2} = \langle v, F'(u)^*w \rangle_{\mathcal{H}},
\end{equation}
analogous to a transpose for function spaces. The adjoint is needed because $F(u) \in L^2$, but the functional gradient must live in $\mathcal{H}$, the space in which $u$ itself resides.

Equation~\eqref{eq:functional_gradient_closed_form} reveals that the functional gradient is obtained by applying the adjoint of the linearized operator to the residual, $F(u)$. As a result, errors in one region or frequency of the solution propagate globally through $F'(u)^*$.

For nonlinear operators, the linearization itself depends on $u$, so the curvature of the objective evolves during training. For linear PDEs, $F'(u)$ is constant, independent of $u$; for nonlinear PDEs, $F'(u)$ changes as $u$ changes during training. For example,
\begin{equation}
    F(u) = u^3 \quad \Longrightarrow \quad F'(u)[h] = 3u^2h,
\end{equation}
so at $u=2$, the gradient scaling is roughly $12$, while at $u=10$, it is roughly $300$ -- a $25\times$ difference in steepness. This evolving curvature means the optimization landscape itself shifts as training progresses, much like navigating terrain where the hills keep moving.

This evolving curvature is compounded by poor conditioning in multi-scale problems. The condition number
\begin{equation}
    \chi = \lambda_{\max}/\lambda_{\min},
\end{equation}
the ratio of the largest to smallest Hessian eigenvalues captures how uneven this curvature is. A large network learning
\begin{equation}
    u(x) = \sin(x) + \sin(100x)
\end{equation}
simultaneously must contend with gradients of $O(1)$ for the low-frequency component $\sin(x)$ alongside gradients of $O(10^4)$ for the high-frequency component $\sin(100x)$, yielding $\chi > 10^4$ and forcing tiny learning rates for stable training. Boosting addresses this by separating stages for different scales: each small network is individually well-conditioned, and sequential refinement avoids the need to learn all scales simultaneously within a single model.

As Equation~\eqref{eq:functional_gradient_closed_form} shows, the residual couples all components of the solution, so representing all scales within a single network can induce poor conditioning and unstable gradient dynamics (\cite{wang-2020}).


\paragraph{Contribution} We propose a variational boosting formulation that decomposes the solution into a sequence of additive correction networks. Each stage trains a weak learner whose converged correction satisfies a local orthogonality condition — equivalent to a projected functional gradient descent step onto the tangent space of the network's function manifold (Section~\ref{subsection:projected_descent_and_optimality}). Crucially, because each weak learner is small, each stage admits full second-order optimization, enabling Newton or conjugate gradient (CG) updates (\cite{nocedal-2006}). We use second-order optimization methods for nonstiff ordinary differential equations (ODEs), Sections~\ref{subsubsection:duffing_equation}--\ref{subsubsection:nrd_equation}, because the parameter counts for the weak learners are small for these cases. In all other cases, second-order optimization is omitted. This framework yields 1) a variational interpretation of multi-stage PINNs, 2) a functional gradient perspective, and 3) second-order solvability at each stage.

\paragraph{Related Work} To our knowledge, our approach is the first to provide a theoretical justification for variational boosting in PINNs, supported by experimental results; the first to exploit the small network sizes of the weak learners to enable second-order optimization; and the first to utilize transfer learning between correction stages. Prior work has explored applying boosting to PINNs~\citep{fang-2023}, but did not employ these techniques. Although the examples in the two papers are not directly comparable, for similar examples both our approach and ~\citep{fang-2023}'s approach achieve relative $L^2$ errors of the same order of magnitude. It is worth noting that our correction models are much smaller, uniform in size, do not use Fourier features, and are trained for fewer epochs. 

Compared to other ensemble methods, such as Mixture-of-Experts~\citep{inproceedings}, the MSE achieved in this paper is one order of magnitude lower, despite our correction models being smaller and trained for fewer epochs. The example used in that paper is a linear PDE with an L-shaped domain, which is different than the nonlinear PDEs considered throughout this paper.

\section{Variational Boosting Framework}
\label{section:variation_boosting_framework}

\subsection{Problem Setup}
\label{subsection:problem_setup}

Let $\Omega \subset \mathbb{R}^d$ be a bounded domain with Lipschitz boundary $\partial\Omega$. We seek a solution $u \in H^m(\Omega)$, the Sobolev space of functions with $m$ derivatives in $L^2(\Omega)$. Sobolev spaces form a nested hierarchy,
\begin{equation}
    L^2(\Omega) = H^0(\Omega) \supset H^1(\Omega) \supset H^2(\Omega) \supset \cdots,
\end{equation}
with each successive space imposing an additional derivative-regularity requirement. This regularity is necessary since the PDEs we consider involve derivatives of $u$ up to order $m$. Both $L^2(\Omega)$ and $H^m(\Omega)$ are Hilbert spaces under their respective inner products, a property used throughout this section (e.g., reflexivity in the existence argument of Appendix~\ref{appendix:conditions_existence}).

We consider a nonlinear differential operator $F \colon H^m(\Omega) \to L^2(\Omega)$ of the form
\begin{equation}
    F(u) = Du + g(u) + f(x).
\end{equation}
Here $D$ is a linear differential operator of order $m$ that maps $H^m(\Omega)$ continuously into $L^2(\Omega)$, $g \colon \mathbb{R} \to \mathbb{R}$ is a smooth nonlinearity, and $f \in L^2(\Omega)$ is a prescribed source term. We assume $m > d/2$, so that the Sobolev embedding $H^m(\Omega) \hookrightarrow L^\infty(\Omega)$ holds. Since $\Omega$ has finite measure, this implies $g(u) \in L^\infty(\Omega) \subset L^2(\Omega)$ for all $u \in H^m(\Omega)$, and therefore $F$ is well-defined as a map $H^m(\Omega) \to L^2(\Omega)$.

The space $L^2(\Omega)$ is a Hilbert space with inner product
\begin{equation}
    \langle \phi, \psi \rangle_{L^2(\Omega)}
    = \int_{\Omega} \phi(x)\,\psi(x)\,dx,
\end{equation}
and associated norm
\begin{equation}
    \|\phi\|_{L^2(\Omega)}
    = \left(\int_{\Omega} |\phi(x)|^2\,dx\right)^{\!1/2}.
\end{equation}
We seek a real-valued solution $u \in H^m(\Omega)$ of the strong-form PDE
\begin{equation}
    F(u)(x) = 0 \quad \text{for almost everywhere } x \in \Omega,
\end{equation}
subject to appropriate boundary conditions on $\partial\Omega$. For a given approximation $u_\theta$, the quantity $F(u_\theta) \in L^2(\Omega)$ is the PDE residual, and PINN training enforces $F(u_\theta)\approx 0$ by minimizing an empirical approximation of $\|F(u_\theta)\|_{L^2(\Omega)}^2$.

We work in $\mathcal{H} = H^m(\Omega)$, equipped with inner product $\langle \cdot, \cdot \rangle_{\mathcal{H}}$ and corresponding norm $\|\cdot\|_{\mathcal{H}}$. We take $\langle \cdot,\cdot \rangle_{\mathcal{H}}$ to be the standard $H^m(\Omega)$ inner product,
\begin{equation}
    \langle u, v \rangle_{\mathcal{H}} = \sum_{|\alpha| \leq m} \langle D^\alpha u,\, D^\alpha v \rangle_{L^2(\Omega)},
\end{equation}
with associated norm $\|u\|_{\mathcal{H}} = \langle u, u \rangle_{\mathcal{H}}^{1/2}$, rather than a residual- or operator-induced seminorm. This choice fixes the Riesz map used to compute the functional gradient $\nabla_{\mathcal{H}}\mathcal{L}(u)$ below, and is distinct from the $L^2(\Omega)$ pairing used in the monotonicity condition~(\ref{equation:monotonicity}).

The governing equation is $F(u)=0$ in $\Omega$, subject to boundary conditions $B(u)=0$ on $\partial\Omega$, where $B$ is a boundary operator. We assume $m$ is large enough that $B(u)$ is well-defined and satisfies $B(u)\in L^2(\partial\Omega)$ for all $u \in \mathcal{H}$ (\cite{adams_fournier}). Following standard PINN practice, boundary conditions are enforced softly by augmenting the loss functional with a weighted penalty term:
\begin{equation}
\mathcal{L}(u) = \|F(u)\|^2_{L^2(\Omega)} + \zeta \,\|B(u)\|^2_{L^2(\partial\Omega)},
\label{eq:weighted_loss}
\end{equation}
where $\zeta> 0$ is a penalty weight. Equation~(\ref{eq:weighted_loss}) defines the scalar loss minimized by the PINN. The associated (idealized) variational problem is
\begin{equation}
    u^* = \underset{u \in \mathcal{H}}{\arg\min}\ \mathcal{L}(u).
\end{equation}
Because boundary conditions are enforced softly, the minimizer $u^*$ satisfies $B(u^*)=0$ only approximately, with accuracy depending on the penalty weight $\zeta$. The interior residual term $\|F(u)\|^2_{L^2(\Omega)}$ is the primary object of our theoretical analysis.

For simplicity, we drop the boundary penalty term in the analysis and, from now on, let
\[
\mathcal L(u) := \|F(u)\|^2_{L^2(\Omega)}.
\]
Note that in practice we train the full weighted loss \eqref{eq:weighted_loss}; omitting the boundary term here is for notational simplicity. Since $\mathcal H=H^m(\Omega)\hookrightarrow L^2(\Omega)$, we use the $L^2(\Omega)$ inner product to pair $F(u)-F(v)\in L^2(\Omega)$ with $u-v\in\mathcal H\subset L^2(\Omega)$. Assume that $F$ is strongly monotone on $\mathcal H$, i.e.,
\begin{equation}
    \langle F(u) - F(v),\, u - v \rangle_{L^2(\Omega)}
    \;\geq\; \gamma\,\|u - v\|_{\mathcal{H}}^2, \quad \gamma > 0,
    \label{equation:monotonicity}
\end{equation}
for all $u, v \in \mathcal{H}$. Under this condition, the strong-form equation $F(u)=0$ has at most one solution in $\mathcal H$ (see Appendix~\ref{appendix:conditions_uniqueness} for the proof).

In the boosted PINN, each stage $k$ solves a minimization subproblem over a hypothesis class $\mathcal U_k \subset \mathcal{H}$. We assume that each stagewise subproblem admits a unique minimizer. This uniqueness assumption ensures that every stage of the boosted PINN is well-defined.

\subsection{Gradient Descent in Function Space}
\label{subsection:gradient_descent_in_function_space}
To motivate the boosted PINN, we present the idealized gradient descent iteration in $\mathcal{H}$. We consider
\[
\mathcal L(u) := \|F(u)\|^2_{L^2(\Omega)} .
\]
(The extension to the full weighted loss \eqref{eq:weighted_loss} is analogous.) The Fr\'echet derivative of $\mathcal L$ at $u$ in the direction $v \in \mathcal{H}$ is
\begin{equation}
    D_v\mathcal{L}(u)
    = 2\,\bigl\langle F(u),\, F'(u)[v] \bigr\rangle_{L^2(\Omega)},
    \label{eq:frechet_derivative}
\end{equation}
where the linearized operator $F'(u) \colon \mathcal{H} \to L^2(\Omega)$ acts as
\begin{equation}
    F'(u)[v] = Dv + g'(u)\,v.
\end{equation}
By the Riesz representation theorem, there exists a unique element
$\nabla_{\mathcal{H}}\mathcal{L}(u) \in \mathcal{H}$ (the functional gradient) satisfying
\begin{equation}
    D_v\mathcal{L}(u)
    = \bigl\langle \nabla_{\mathcal{H}}\mathcal{L}(u),\, v \bigr\rangle_{\mathcal{H}}.
\end{equation}
The idealized gradient descent iteration in $\mathcal{H}$ then reads
\begin{equation}
    u_{k+1} = u_k - \alpha\,\nabla_{\mathcal{H}}\mathcal{L}(u_k)
            = u_k - 2\alpha\,F'(u_k)^{*}F(u_k),
    \label{eq:functional_gd}
\end{equation}
where $\alpha > 0$ is a step size and $k$ is an iteration index. Each update subtracts a correction aligned with the adjoint-weighted residual, driving $F(u_k)$ toward zero.

This iteration is intractable directly for three reasons. First, $u_k$ lives in the infinite-dimensional space $\mathcal{H}$, so the update cannot be represented or stored exactly. Second, the adjoint $F'(u_k)^*$ is itself a differential operator with no general closed form. Third, when the Hessian-like operator $F'(u_k)^*F'(u_k)$ is ill-conditioned (e.g., in stiff problems where eigenvalues span many orders of magnitude), the step size $\alpha$ must be taken prohibitively small to ensure stability, severely slowing convergence.

The boosted PINN replaces each intractable functional update \eqref{eq:functional_gd} with a finite-dimensional correction fit by a neural network. At stage $k$, a network $h_k$ is trained to approximate the functional gradient direction $-2F'(u)^*F(u)$ at the current iterate, and the solution is updated as $u^{(k)} = u^{(k-1)} + h_k$, with $h_k$ chosen to maximally reduce $\mathcal{L}$. 

This transforms the intractable infinite-dimensional iteration into a sequence of tractable finite-dimensional regression problems, each justified by the existence and uniqueness guarantees established in Section~\ref{subsection:problem_setup}.

\subsection{Projected Descent and Optimality}
\label{subsection:projected_descent_and_optimality}
This section illustrates the boosting update rule by showing that each stage solves a restricted minimization in the neural network function class $\mathcal{U}_k$, that the first-order optimality condition forces the residual gradient to be orthogonal to the tangent space of $\mathcal{U}_k$ at $h_k$ after the update, and that the correction $h_k$ can therefore be interpreted as a projected gradient descent step in function space. This connects the practical boosting procedure back to the idealized functional gradient descent iteration from Section~\ref{subsection:gradient_descent_in_function_space}.

The boosted PINN restricts the search at each stage $k$ to a function class $\mathcal{U}_k \subset \mathcal{H}$, taken to be the set of functions representable by the $k$-th neural network (Section~\ref{subsection:problem_setup}). Concretely, let $\phi_k(\cdot;\theta_k)$ denote the $k$-th network (as a function of its input), and define
\[
\mathcal{U}_k := \{\phi_k(\cdot;\theta) : \theta \in \Theta_k\} \subset \mathcal{H}.
\]

At stage $k$, we solve the restricted problem
\begin{equation}
    h_k = \underset{h \in \mathcal{U}_k}{\arg\min}\;
    \mathcal{L}(u^{(k-1)} + h).
    \label{eq:restricted_problem}
\end{equation}
In practice, $h = \phi_k(\cdot;\theta)$ for some network weights $\theta\in\Theta_k$. The optimization searches over all such $h$ to find the one that, when added to the current solution, minimizes the loss. The accumulated solution is then
\begin{equation}
    u^{(k)} = u^{(k-1)} + h_k.
    \label{eq:update_rule}
\end{equation}
Since $\mathcal{U}_k$ is a nonlinear, non-convex set, the first-order optimality condition requires that the directional derivative vanish along all tangent directions $\delta h \in T_{h_k}\mathcal{U}_k$ at $h_k$:
\begin{equation}
    \bigl\langle \nabla_{\mathcal{H}} \mathcal{L}(u^{(k)}),\, \delta h
    \bigr\rangle_{\mathcal{H}} = 0
    \quad \text{for all } \delta h \in T_{h_k}\mathcal{U}_k,
    \label{eq:orthogonality_tangent_space}
\end{equation}
where $\delta h$ is the infinitesimal function perturbation in $T_{h_k}\mathcal{U}_k$. Equivalently, (\ref{eq:orthogonality_tangent_space}) can be expressed as
\begin{equation}
    \nabla_{\mathcal{H}} \mathcal{L}(u^{(k)}) \perp T_{h_k}\mathcal{U}_k.
    \label{eq:orthogonality_tangent_space_2}
\end{equation}
This means that after adding the correction $h_k$, the remaining functional gradient is orthogonal to the tangent space of $\mathcal{U}_k$ at $h_k$. No further descent within $\mathcal{U}_k$ is possible from this point. 

We emphasize that $\mathcal{U}_k$ is the full nonlinear range of the $k$-th network architecture over its entire parameter space $\Theta_k$: warm-starting each weak learner from the previous stage's converged weights (Section~\ref{subsection:boosted_pinn_framework}) serves only to initialize the optimization and does not restrict $\Theta_k$. All parameters of $\phi_k(\cdot;\theta)$ remain free during training. Consequently, $\mathcal{U}_k$ is not an affine subspace of $\mathcal{H}$; it is a nonlinear, non-convex manifold, and the affine structure appearing in the optimality condition~(\ref{eq:orthogonality_tangent_space}) is strictly local, arising from the tangent space $T_{h_k}\mathcal{U}_k$ at the trained weights $\theta_k$. Concretely, this tangent space is spanned by the network's parameter Jacobian at $\theta_k$,
\begin{equation}
    T_{h_k}\mathcal{U}_k = \operatorname{span}\left\{ \partial_{\theta_i}\phi_k(\cdot;\theta) \Big|_{\theta = \theta_k} \;:\; i = 1,\dots,|\theta_k| \right\} \subset \mathcal{H}.
\end{equation}
It is the linear space of directions reachable by an infinitesimal change in the network's weights. The projection interpretation given below should therefore be read as a first-order, local statement about this tangent space, rather than a global affine restriction of the search space $\mathcal{U}_k$ itself.

The update $h_k$ therefore extracts all available descent from the current network class, and can be interpreted as a projected gradient descent step in function space: rather than moving in the full steepest descent direction $-\nabla_{\mathcal{H}}\mathcal{L}(u^{(k-1)})$, one moves in its projection onto the tangent space of $\mathcal{U}_k$:
\begin{equation}
    h_k \approx -\operatorname{proj}_{T_{h_k}\mathcal{U}_k}
    \nabla_{\mathcal{H}}\mathcal{L}(u^{(k-1)}).
\end{equation}
It is an approximation because it is a first-order, intuitive description of what $h_k$ does, not an exact characterization. The exact characterization is (\ref{eq:restricted_problem}). Each subsequent boosting stage introduces a new function class $\mathcal{U}_{k+1}$ to continue reducing the residual that remains orthogonal to $T_{h_k}\mathcal{U}_k$.

\subsection{Boosted PINN Framework}
\label{subsection:boosted_pinn_framework}
Now, we describe the boosting algorithm. We construct the solution sequentially as an additive ensemble of neural networks. The number of trainable parameters for a given neural network is denoted as $|\theta|$. The first stage, hereafter called \emph{stage 0}, is a standard PINN trained directly on the full problem, with $h_0 \in \mathcal{U}_0$ given by
\begin{equation}
    h_0 = \underset{h \in \mathcal{U}_0}{\arg\min}\; 
    \mathcal{L}(h),
\end{equation}
giving the initial approximation $u^{(0)} = h_0$. This is consistent with the general update rule \eqref{eq:update_rule} by setting $u^{(-1)} = 0$. The base learner $h_0$ carries no shrinkage parameter since it is trained directly on the full problem with no prior approximation to correct. For each subsequent boosting stage $k \geq 1$, given the current approximation $u^{(k-1)}$, we train a new weak learner $h_k \in \mathcal{U}_k$ to reduce the residual of the accumulated model, as given by (\ref{eq:restricted_problem}). We update the ensemble:
\begin{equation}
    u^{(k)} = u^{(k-1)} + \alpha_k h_k.
\end{equation}
Here, $\alpha_k > 0$ is a shrinkage parameter controlling the contribution of each new learner. The cumulative approximation after $K$ boosting stages is therefore:
\begin{equation}
    u^{(K)} = h_0 + \sum_{k=1}^{K} \alpha_k h_k,
\end{equation}
making explicit that $h_0$ is the stage 0 weak learner and each subsequent $h_k$ corrects the residual left by the previous ensemble.
To see why this is a gradient boosting procedure in function space, we 
linearize $F$ around $u^{(k-1)}$ for small $h$:
\begin{equation}
    F(u^{(k-1)} + h) \approx r^{(k-1)} + F'(u^{(k-1)})[h],
\end{equation}
where $r^{(k-1)} = F(u^{(k-1)}) \in L^2(\Omega)$ is the current residual and $F'(u^{(k-1)}) \colon \mathcal{H} \to L^2(\Omega)$ is the linearized operator. The Fréchet derivative of $\mathcal{L}$ at $u^{(k-1)}$ in direction $h$ is:
\begin{equation}
    D_h\mathcal{L}(u^{(k-1)}) 
    = 2\,\bigl\langle r^{(k-1)},\, F'(u^{(k-1)})[h] \bigr\rangle_{L^2(\Omega)} 
    = 2\,\bigl\langle F'(u^{(k-1)})^*\, r^{(k-1)},\, h \bigr\rangle_{\mathcal{H}},
\end{equation}
so the steepest descent direction at $u^{(k-1)}$ is:
\begin{equation}
    -\nabla_{\mathcal{H}}\mathcal{L}(u^{(k-1)}) 
    = -2\,F'(u^{(k-1)})^*\, r^{(k-1)},
\end{equation}
where $F'(u^{(k-1)})^* \colon L^2(\Omega) \to \mathcal{H}$ is the adjoint of the linearized operator. Each new weak learner $h_k$ is trained to minimize the residual of the updated ensemble; by the above linearization, this is \emph{approximately} equivalent to finding the element of $\mathcal{U}_k$ that best approximates the projection of $-\nabla_{\mathcal{H}}\mathcal{L}(u^{(k-1)})$ onto $T_{h_k}\mathcal{U}_k$.

\subsubsection{Linearization of Residual Operator $F$}

We consider two variants of the boosted PINN correction step, distinguished by whether the residual operator $F$ is linearized before training $h_k$. 

In the linearized variant, we freeze $F$ at $u^{(k-1)}$ via its first-order expansion,
\begin{equation}
    F(u^{(k-1)} + \alpha h) \approx r^{(k-1)} + \alpha\, F'(u^{(k-1)})\, h,
\end{equation}
and train $\theta_k$ to minimize this fixed linear model of the residual. This is a Gauss--Newton subproblem. Even though we solve it with Adam rather than a closed-form linear solve (since $h_k = \phi_k(\cdot;\theta_k)$ is nonlinear in its parameters), the objective itself only ever sees the residual's local linear approximation at the start of the stage. As a result, training approaches the Gauss--Newton solution rather than a solution of the true nonlinear problem. 

In the full nonlinear variant, we instead train $\theta_k$ directly against the true, unlinearized residual $\|F(u^{(k-1)} + \alpha h_k)\|^2$, re-evaluating $F$ (and its derivative, via autodiff) at the moving iterate $u^{(k-1)} + \alpha h_k$ throughout training rather than only at the starting point. The two variants therefore differ not in the optimizer, but in the objective being optimized. The linearized variant solves a single frozen linear model, with error determined by how far $h_k$ departs from the point of linearization. The full nonlinear variant follows the true nonlinear landscape at every step. The drawback is that there is no longer a clean closed-form characterization of what the converged $h_k$ represents. 

The decision to linearize $F$ or not was determined via experimentation. The method that resulted in the lowest MSE was chosen. The type of loss used is shown in Table \ref{tab:ode_overview} for the ODE examples and Table \ref{tab:pde_examples} for the PDE examples.

\section{Model Properties}
\label{section:optimization_properties}

\subsection{Monotone Decrease}
\begin{figure}[ht]
    \centering
    \includegraphics[width=1.0\textwidth]{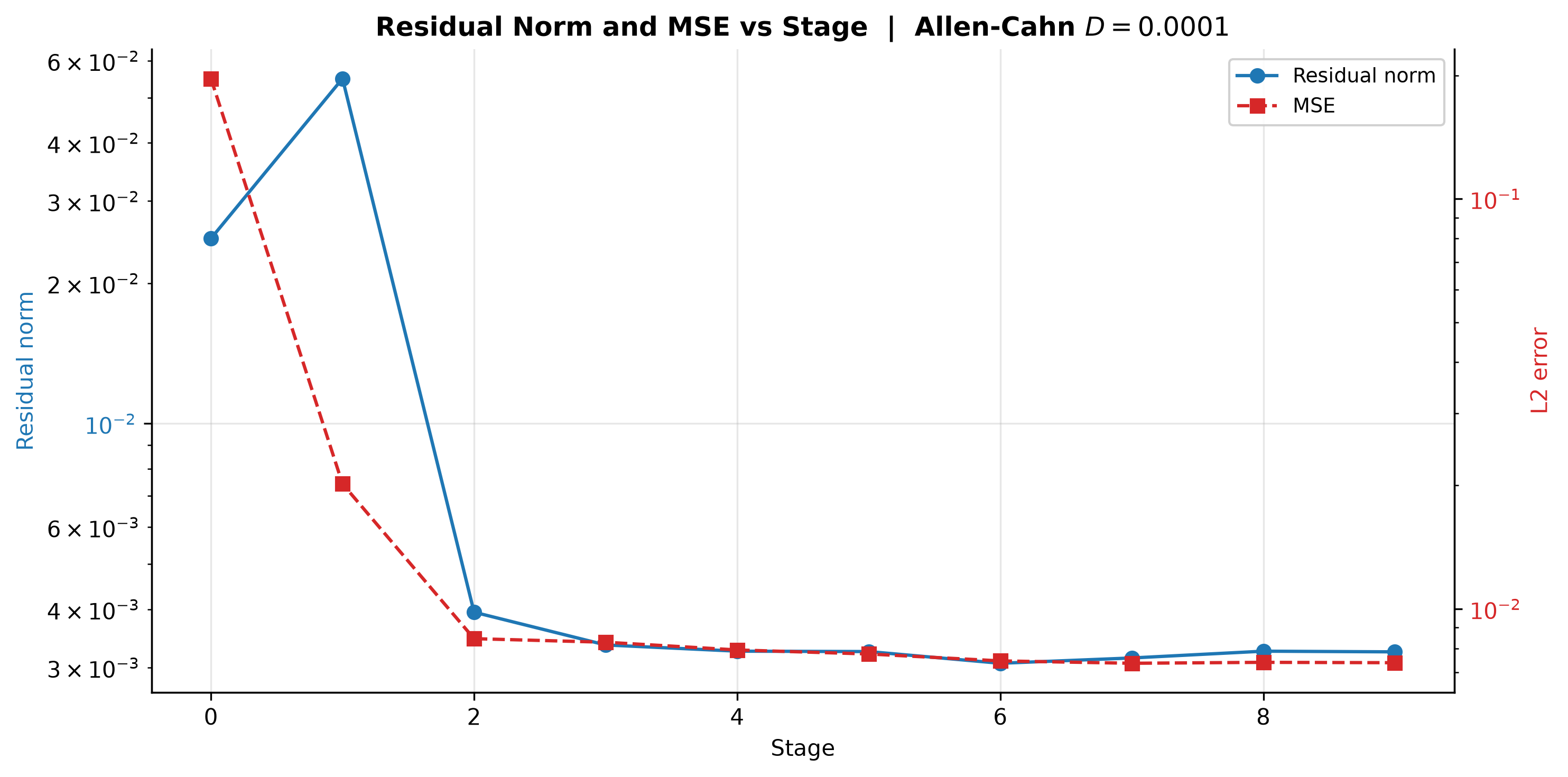}
    \caption{Boosted PINN metrics per stage.}
    \label{fig:allencahn_periodic_loss_monotone}
\end{figure}

Assuming $0 \in \mathcal{U}_k$ is always a feasible choice (e.g., by including a zero-initialization), the exact minimizer $h_k$ satisfies
\begin{equation}
    \mathcal{L}(u^{(k-1)} + h_k) 
    \leq \mathcal{L}(u^{(k-1)} + 0) 
    = \mathcal{L}(u^{(k-1)}).
\end{equation}
Thus, provided each stage solves its restricted problem exactly and $\alpha_k = 1$, the loss sequence $\{\mathcal{L}(u^{(k)})\}_{k \geq 0}$ is non-increasing:
\begin{equation}
    \mathcal{L}(u^{(0)}) \geq \mathcal{L}(u^{(1)}) \geq \mathcal{L}(u^{(2)}) 
    \geq \cdots \geq 0.
\end{equation}
In practice, two factors break this guarantee. First, $\alpha_k \in (0,1)$ is a fixed shrinkage parameter rather than an exact line search, so the update is not guaranteed to be the exact minimizer assumed above. Second, our Newton-based optimizer rolls back to the best result obtained during the preceding Adam phase, rather than to the idealized fallback $h_k = 0$; as a result, monotone decrease relative to $u^{(k-1)}$ is not strictly guaranteed. Nevertheless, we observe an overall decreasing loss trend across stages in practice; Figure~\ref{fig:allencahn_periodic_loss_monotone} illustrates this for the stiff Allen--Cahn (periodic) example, where the loss decreases approximately monotonically despite the absence of a rollback to $h_k = 0$.

\subsection{Model Assumptions}
\label{section:model_assumptions}
Our model assumptions are as follows. Mathematically, the solution is sufficiently smooth for automatic differentiation to evaluate the residual. The residual operator $F$ is coercive and strongly monotone (Section~\ref{subsection:problem_setup}), ensuring the PDE has exactly one solution in $\mathcal{H}$. Each function class $\mathcal{U}_k$ contains the zero function, i.e., $0 \in \mathcal{U}_k$. In the idealized setting described in Sections~\ref{section:variation_boosting_framework}--\ref{section:optimization_properties}, each correction stage reduces the residual. In practice, this is not strictly guaranteed due to fixed shrinkage and the optimizer's rollback behavior, though we observe an overall decreasing trend empirically.

For the algorithm to hold, we assume the residual at collocation points is a sufficient proxy for the continuous residual. The weak learners have sufficient capacity to approximate the correction at each stage. The shrinkage coefficients $\alpha_k \in (0,1)$ are chosen small enough that the loss does not oscillate or diverge. A large $\alpha_k$ can cause the ensemble to overfit the residual at a single stage, destabilizing subsequent stages.

\subsection{Model Complexity}

\paragraph{Inference Time Complexity} For a fully connected network with width $w$, depth $L$, and $N$ collocation points, the forward pass cost is computed as follows. Each layer is a matrix-vector product of size $w \times w$, so the cost is $O(w^2)$. Over $L$ layers, we get $O(w^2 L)$. Over $N$ collocation points, we get $O(N w^2 L)$. Finally, over $K$ stages, we arrive at $O(KN w^2 L)$. Note that this computation assumes uniform width across layers, which holds for all networks in this paper (Tables~\ref{tab:architecture_boosted} and \ref{tab:architecture_std}).
Let $W$ denote the width of the standard PINN. Using the same logic, we arrive at $O(N W^2 L)$. Therefore, assuming equal depth $L$ (as with all examples in this paper), collocation points $N$, and number of training iterations, the boosted PINN is more efficient if $Kw^2 \ll W^2$.

\paragraph{Space Complexity} Each weak learner requires $O(w^2 L)$ parameters. Storing all $K$ stages simultaneously requires $O(K w^2 L)$ parameters. Therefore, the space complexity scales linearly with the number of stages.

\paragraph{Sample Size} In our experiments, we have two approaches. The first approach uses the same $N$ points across all stages, giving a total sample size of $N$. The second, used in the Allen--Cahn experiments, resamples $N$ fresh points at each stage, giving total sample size $KN$ (see Appendix~\ref{section:sampling_strategy} for further details).

\section{Second-Order Optimization}
\label{sec:second_order_optimization}

Because $\mathcal{U}_k$ is represented by a small neural network, the restricted problem is low-dimensional. A second-order solver is computationally feasible at each stage, unlike for monolithic PINNs.

\subsection{Conjugate Gradient}
\label{subsection:conjugate_gradient}
Let $H$ denote the Hessian matrix of the loss $\mathcal{L}(\theta) := \mathcal{L}(\phi_k(\cdot;\theta))$ with respect to the parameters $\theta$. $H$ is not guaranteed to be positive-definite in practice, so we apply Tikhonov regularization and solve the linear system
\begin{equation}
    (H + \gamma I) d = -g,
\end{equation}
where $g = \nabla_\theta \mathcal{L}(\theta)$ is the gradient, $\gamma > 0$ is a small scalar, and $d$ is the Newton direction. The parameter update is then
\begin{equation}
    \theta_{new} = \theta_{old} + \tau_{ls}\, d,
\end{equation}
where $\tau_{ls}$ is the step size along direction $d$.

Conjugate gradient (CG) is used to solve this system without explicitly forming $H$. Instead, CG requires only Hessian-vector products $Hv$. For a fixed vector $v$, this product can be computed via a second-order directional derivative:
\begin{equation}
    Hv = \nabla_\theta\bigl(\nabla_\theta \mathcal{L}(\theta) \cdot v\bigr),
\end{equation}
since differentiating $\nabla_\theta \mathcal{L}(\theta)$ in the direction of $v$ yields exactly $Hv$. This avoids forming and storing the full Hessian.

The CG algorithm solves $Ax = b$, where $A = H + \gamma I$ and $b = -g$; upon convergence, the solution $x$ gives the Newton direction $d$. The initial residual of the linear system is $\tilde{r}_0 = b - Ax_0$. Note the tilde to distinguish this from the PDE residual in Section~\ref{subsection:boosted_pinn_framework}. Initialize $x_0 = 0$, so that
\begin{equation}
    p_0 = \tilde{r}_0 = b = -g,
\end{equation}
meaning the first search direction is the steepest descent direction.

At each subsequent iteration $k$, compute the step size
\begin{equation}
    \tau_k = \frac{\tilde{r}_k^T \tilde{r}_k}{p_k^T A p_k},
\end{equation}
and update the iterate, residual, and search direction:
\begin{equation}
    x_{k+1} = x_k + \tau_k p_k,
\end{equation}
\begin{equation}
    \tilde{r}_{k+1} = \tilde{r}_k - \tau_k A p_k,
\end{equation}
\begin{equation}
    p_{k+1} = \tilde{r}_{k+1} + \beta_k p_k, \qquad 
    \beta_k = \frac{\tilde{r}_{k+1}^T \tilde{r}_{k+1}}
                  {\tilde{r}_k^T \tilde{r}_k}.
\end{equation}
The update for $\beta_k$ is chosen so that all search directions satisfy the conjugacy condition:
\begin{equation}
    p_i^T A p_j = 0, \qquad i \neq j.
\end{equation}

The quantity $p_i^T A p_j$ is a scalar measuring how two directions interact under the curvature defined by $A$. When it is zero, the directions are $A$-orthogonal, meaning progress along one direction does not interfere with progress along another. One can think of it as a curvature-weighted inner product between two directions. At each step, the algorithm incorporates new curvature information and eliminates error along a new axis, converging in at most $n$ steps for an $n \times n$ system in exact arithmetic. In practice, finite precision may require additional iterations.

\subsection{Newton's Method}
\label{subsection:newtons_method}
Compared to the conjugate gradient method, Newton's method computes the entire Hessian explicitly, making it feasible only for small models. The Newton system is:
\begin{equation}
    H d = -g,
\end{equation}
where $g = \nabla_\theta \mathcal{L}(\theta)$ is the gradient, 
$H = \nabla^2_\theta \mathcal{L}(\theta)$ is the Hessian, and $d$ is the Newton direction. The parameter update is then:
\begin{equation}
    \theta_{k+1} = \theta_k + d.
\end{equation}

\paragraph{Hessian Computation}
In the code, $H$ is computed column by column using the identity:
\begin{equation}
    H\vec{e}_i = \nabla_{\theta}(\vec{e}_i^T \nabla_{\theta}
    \mathcal{L}(\theta)),
\end{equation}
where $\vec{e}_i$ is the $i$-th standard basis vector. This is slow because it requires one backward pass per column of the Hessian and keeps a full computation graph in memory for second derivatives.

\paragraph{Tikhonov Regularization}
$H$ is not guaranteed to be positive-definite in practice. After the Hessian has been computed, if the Hessian has small or negative 
eigenvalues, Tikhonov regularization is applied:
\begin{equation}
    H_{\text{reg}} = H + \gamma I,
\end{equation}
where $\gamma > 0$ is by default set to $10^{-3}$. We then check the minimum eigenvalue of $H_{\text{reg}}$. If $\lambda_{\min}(H) < 0$, the regularization is strengthened:
\begin{equation}
    H_{\text{reg}} = H + (\gamma + |\lambda_{\min}| + \epsilon)\,I,
\end{equation}
where $\epsilon = 10^{-6}$ ensures strict positive-definiteness.
\paragraph{Trust Region}
To ensure the step size is not too large, we enforce a trust region 
constraint:
\begin{equation}
    \|d\|_2 \leq \delta,
\end{equation}
where $\delta$ is the trust region radius, initialized to $1$. If 
$\|d\|_2 > \delta$, the update is rescaled to 
$d \leftarrow \delta \cdot d / \|d\|_2$.
After applying the update, the trust region radius is adjusted using the trust ratio:
\begin{equation}
    \rho = \frac{\text{actual reduction}}{\text{predicted reduction}},
\end{equation}
where the actual reduction is how much the loss decreased:
\begin{equation}
    \text{actual reduction} = \mathcal{L}(\theta_{old}) 
    - \mathcal{L}(\theta_{new}),
\end{equation}
and the predicted reduction is given by the quadratic Taylor model 
\begin{equation}
    m(d) = \mathcal{L}(\theta) + g^T d + \frac{1}{2}d^T H d.
\end{equation}
So then,
\begin{equation}
    \text{predicted reduction} = \mathcal{L}(\theta) - m(d) 
    = -(g^T d + \tfrac{1}{2}d^T H d).
\end{equation}
Every computed step is applied; $\rho$ is used only to adjust the trust region radius for the subsequent iteration:
\begin{itemize}
    \item $\rho < 0.25$: poor agreement with the quadratic model; 
    shrink $\delta \leftarrow 0.5\delta$.
    \item $0.25 \leq \rho \leq 0.75$: acceptable agreement; keep 
    $\delta$ unchanged.
    \item $\rho > 0.75$ \emph{and} $\|d\|_2 \geq 0.9\,\delta$: good 
    agreement and the step was near the trust boundary; expand 
    $\delta \leftarrow 2\delta$.
\end{itemize}

\paragraph{Stopping Conditions}
Inspired by common implementations of quasi-Newton methods such as L-BFGS, we employ several stopping criteria. The optimization is terminated when any of the following conditions is satisfied:
\begin{enumerate}
    \item \textbf{Maximum number of iterations.} The number of optimization steps reaches a prescribed upper bound (typically $10$--$20$ iterations).
    \item \textbf{Gradient norm threshold.} The Euclidean norm of the gradient falls below a specified tolerance, indicating proximity to a local optimum:
    \[
        \|\nabla \mathcal{L}(\theta)\|_2 < \varepsilon_{\text{grad}}.
    \]
    \item \textbf{Parameter update threshold.} The norm of the parameter update becomes sufficiently small, suggesting further updates will have a negligible effect:
    \[
        \|\Delta \theta\|_2 < \varepsilon_{\text{param}}.
    \]
    \item \textbf{Loss change threshold.} The absolute change in the loss between successive iterations falls below a given 
    tolerance:
    \[
        |\mathcal{L}(\theta_k) - \mathcal{L}(\theta_{k-1})| 
        < \varepsilon_{\text{loss}}.
    \]
    \item \textbf{Relative loss change threshold.} The loss change relative to the magnitude of the previous loss falls below a given tolerance:
    \[
        \frac{|\mathcal{L}(\theta_k) - \mathcal{L}(\theta_{k-1})|}
        {1 + |\mathcal{L}(\theta_{k-1})|} < \varepsilon_{\text{tol}}.
    \]
\end{enumerate}
\section{Experiments}
\label{sec:experiments} 

For the remainder of this paper, we refer to the baseline model as the \emph{standard PINN}, also referred to as \emph{monolithic PINN} in Sections~\ref{section:variation_boosting_framework}--\ref{sec:second_order_optimization}. All reported performance metrics are averaged over a minimum of 10 independent seeds. We consider a PINN \textit{converged} when the MSE between the PINN solution and the numerical solution satisfies MSE $< 10^{-2}$. Throughout this paper, we follow the PINN convention and refer to the RMSE as MSE. The definitions for each performance metric can be found in Appendix~\ref{appendix:error_residual_metrics}. We use two convergence metrics reported as "Training Time" and "Iterations" in the tables. The former measures the time it takes for the PINN to achieve MSE $< 10^{-2}$, and the latter measures the epochs or function calls (in the case of L-BFGS) required to achieve MSE $< 10^{-2}$. For nonstiff equations, we report results with second-order optimization methods. Otherwise, only the first-order optimization results are shown. 

For the convergence metrics, we say a model does not converge if any single seed (out of 10) fails to converge. The performance metrics are averaged over nonconvergent and convergent training runs. However, we did not observe any nonconvergence for the boosted PINN in any of the following examples. 

To make results comparable, in all the ODE examples, we set the batch size equal to the training set. In other words, the standard PINN and each stage use full-batch gradient descent. The number of epochs for both models is fixed and equal. If the standard PINN trains for $E$ epochs, the boosted PINN has a budget of $E$ epochs, such that $\sum_{i=1}^k E_{weak}^{(k)} = E$, where $E_{weak}^{(k)}$ is the number of epochs used to train the weak learner at stage $k$.

For parameter parity, the architecture of the weak learner is identical to the standard PINN. Since the boosted model is conceptually a single model whose output is the sum of its weak learners, matching each weak learner's architecture to the standard PINN places both models on equal architectural footing, differing only in how their shared capacity is optimized: jointly, in a single training run, versus sequentially, one stage at a time. This isolates the effect of the staged training procedure itself as the sole variable, rather than attributing any performance difference to a difference in model size or capacity.

Lastly, each weak learner is a separate neural network. We use \textit{stage} to refer to the process of training one weak learner. Model architectures for standard and boosted PINNs can be found in Appendix Tables \ref{tab:architecture_boosted} and \ref{tab:architecture_std}, along with experimental setup information.

\subsection{ODE Examples}
We evaluate the boosted PINN on the following nonlinear ODEs: Duffing equation, Van der Pol equation, and a nonlinear reaction–diffusion (NRD) equation. We also include an example on Lotka--Volterra, a coupled ODE system. The NRD equation is a boundary value problem (BVP); the rest are initial value problems (IVPs). All examples are nonlinear. For the Van der Pol equation and the nonlinear reaction–diffusion equation, we show both a stiff and a nonstiff regime by adjusting parameters in each respective equation. Duffing is evaluated on 10 sets of parameters, each with qualitative differences (see Table \ref{tab:duffing_param_sets} for details). Lotka--Volterra is evaluated in a single (nonstiff) regime. For the Van der Pol equation, we show results for $\mu \in \{2.0, 3.0, 4.0\}$. For the NRD equation, we show results for $\kappa \in \{10, 100\}$. The set of ODE examples is summarized in Table~\ref{tab:ode_overview}.

\begin{table}[h]
    \centering
    \begin{tabular}{llll}
        \toprule
        Type & Equation & Parameter Regime & Loss Linearized \\
        \midrule
        Nonlinear IVP                       & Duffing              & 10 parameter sets              & No  \\
        Nonlinear BVP                       & NRD & $\kappa = 10$          & Yes \\
        Stiff Nonlinear BVP                 & NRD & $\kappa = 100$         & Yes \\
        Stiff Nonlinear Reaction Equations  & Van der Pol          & $\mu \in \{2.0, 3.0, 4.0\}$     & Yes \\
        Nonlinear Coupled IVP               & Lotka--Volterra      &                                 & No  \\
        \bottomrule
    \end{tabular}
    \caption{Overview of ODE examples.}
    \label{tab:ode_overview}
\end{table}

\paragraph{Transfer Learning for ODE Examples}
At each boosting stage $k \geq 1$, the network $h_k$ is initialized by transferring weights from the previous stage's network $h_{k-1}$. The pretrained weights are loaded directly, after which the final two layers are re-initialized with Xavier initialization~\citep{pmlr-v9-glorot10a} and rescaled by a stage-specific scaling factor $\xi_k$, detailed in Table~\ref{tab:ode_transfer_learning}. A small $\xi_k$ causes the final layers to start nearly zeroed out; the pretrained layers initially dominate the network's output, and the correction $h_k$ begins as a small perturbation to $u^{(k-1)}$. A larger $\xi_k$ preserves more of the Xavier magnitude in the final layers, introducing greater weight diversity and allowing faster adaptation to the remaining residual. For all experiments, $\xi_k$ is fixed across stages.
As shown in Table~\ref{tab:ablation_transfer}, transfer learning improves the convergence of the boosted PINN, with the magnitude of the effect varying across problems, from negligible for the Duffing oscillator to essential for convergence in the Lotka--Volterra system. In all cases, removing transfer learning increases MSE, though the size of this effect varies substantially by problem.
\begin{table}[h]
    \centering
    \begin{tabular}{lll}
        \toprule
        Type & Name & Scale Factor ($\xi$) \\
        \midrule
        Nonlinear IVP         & Duffing        & 0.5   \\
        Stiff and Nonstiff Nonlinear BVP & NRD & 0.01  \\
        Stiff Nonlinear IVP   & Van der Pol    & 0.05  \\
        Coupled Nonlinear IVP & Lotka--Volterra & 0.001 \\
        \bottomrule
    \end{tabular}
    \caption{Scale factors for ODE examples.}
    \label{tab:ode_transfer_learning}
\end{table}

\subsubsection{Duffing Equation}
\label{subsubsection:duffing_equation}

In this subsection, we show results for the Duffing equation, posed as an IVP. The equation is:
\begin{equation}
    \frac{d^2u}{dt^2} + \delta \frac{du}{dt} + \alpha u + \beta u^3 = \gamma \cos(\omega t).
\end{equation}
Five parameters control the behavior of the ODE. $\delta$ controls damping; the larger the value, the stronger the damping. $\alpha$ is the linear stiffness parameter; the larger the value, the stiffer the equation. $\beta$ is the nonlinear stiffness parameter. $\gamma$ is the amplitude of the forcing function, where $\omega$ is the angular frequency.

For our equation specification, $t \in (0.0, 5.0)$, with initial conditions:
$$
u(0) = 0.5, \qquad \left.\frac{du}{dt}\right|_{t=0} = 0.
$$
We compare the performance of the models on 10 different sets of parameters. The results are averaged to yield the values in Table \ref{tab:duffing_standard} and Table \ref{tab:duffing_boosted}. All models are compared against a numerical solution given by a Runge–Kutta 4(5) solver, with a relative tolerance of $10^{-8}$ and an absolute tolerance of $10^{-10}$, using \lstinline[language=Python]{solve_ivp()}.

Each PINN (weak learner and standard PINN) is a fully connected neural network (FCNN). For the Duffing equation, the architecture is $(t)-32-32-(u)$, two hidden layers, 32 neurons each, with $\sin$ activations. Table~\ref{tab:architecture_std} contains all architectures. For the standard PINN, we use two different optimizers, Adam and L-BFGS. Table~\ref{tab:duffing_standard} summarizes the standard PINN performance.

\begin{table}[ht]
    \centering
    \begin{tabular}{lcc}
        \toprule
        & \multicolumn{2}{c}{Monolithic PINN} \\
        \cmidrule(lr){2-3}
        Metric & Adam & L-BFGS \\
        \midrule
        Residual Norm              & $9.92 \times 10^{-3}$ & $3.88 \times 10^{-3}$ \\
        MSE                        & $1.65 \times 10^{-3}$ & $3.77 \times 10^{-4}$ \\
        Relative $L^2$ Error       & $3.40 \times 10^{-3}$ & $6.19 \times 10^{-4}$ \\
        Training Time (s)          & $3.25$                & $0.31$                \\
        Number of Iterations       & $1{,}530$             & $132$                 \\
        Learning Rate              & $10^{-3}$             & $1.0$                 \\
        Collocation Points         & $1{,}000$             & $1{,}000$             \\
        Network Size               & $2{,}209$             & $2{,}209$             \\
        \bottomrule
    \end{tabular}
    \caption{Monolithic PINN performance on the Duffing equation.}
    \label{tab:duffing_standard}
\end{table}

Each weak learner in the boosted PINN is an FCNN. Architectures are fixed across stages. We train the boosted PINN under three optimization schemes: Adam, Adam + CG, and Adam + Newton. When Adam is the sole optimizer, each weak learner is trained only with Adam. When a second-order optimizer is introduced, each boosting stage uses Adam for approximately $90 \pm 5\%$ of epochs before switching to the second-order optimizer for the remaining epochs. The exact number of epochs varies across stages, but this ratio is held constant across stages. 

The rationale for this hybrid strategy is as follows. Since the loss surface for nonlinear problems is nonconvex and noisy, Adam is used first to descend toward a region of the loss surface that is well-conditioned. Second-order methods are sensitive to initialization: if the Hessian $H$ is not positive definite, the optimization step can diverge. By allowing Adam to reach a well-posed region first, the second-order optimizer can then exploit local curvature information more reliably. 

Training time refers to the time it takes for the PINN's MSE to fall below $10^{-2}$. Similarly, iterations refer to the number of epochs required. Note that to make the comparison between Adam and L-BFGS accurate, we report the total number of closure calls by the optimizer for the L-BFGS iterations. Table~\ref{tab:hyperparams_boosted} contains information on epochs per stage for all examples. 

\begin{table}[ht]
    \centering
    \begin{tabular}{lccc}
        \toprule
        & \multicolumn{3}{c}{Boosted PINN} \\
        \cmidrule(lr){2-4}
        Metric & Adam & Adam + CG & Adam + Newton \\
        \midrule
        Residual Norm               & $5.32 \times 10^{-3}$ & $5.43 \times 10^{-3}$ & $3.24 \times 10^{-3}$ \\
        MSE                         & $2.25 \times 10^{-4}$ & $7.08 \times 10^{-4}$ & $1.44 \times 10^{-4}$ \\
        Relative $L^2$ Error        & $3.26 \times 10^{-4}$ & $9.64 \times 10^{-4}$ & $2.17 \times 10^{-4}$ \\
        Training Time (s)           & $8.22$               & $7.03$                & $187.72$              \\
        Number of Iterations        & $702$                 & $738$                 & $407$                 \\
        Learning Rate               & $10^{-2}$             & $10^{-2}$             & $10^{-2}$             \\
        Collocation Points          & $1{,}000$             & $1{,}000$             & $1{,}000$             \\
        Network Size                & $2{,}209$             & $2{,}209$             & $2{,}209$             \\
        \bottomrule
    \end{tabular}
    \caption{Boosted PINN performance on the Duffing equation.}
    \label{tab:duffing_boosted}
\end{table}

The boosted PINN with Newton performs best across all metrics: residual norm, MSE, and relative $L^2$ error. However, it has the slowest convergence time at 187.72 seconds. The Adam + Newton approach is slowest because, to ensure stable training, the Newton method regularizes $H$ and loops over many candidate steps until convergence conditions are satisfied or until max iterations are reached, before taking a Newton step. It is worth noting that the boosted PINN's total training time is, in general, higher than the standard PINN's, since it trains $K$ sequential stages, each requiring a full optimization run.

In terms of convergence, the standard PINN with L-BFGS converges fastest at 0.31 seconds because the parameter sets chosen for the Duffing equation example are smooth, non-chaotic regimes where standard PINNs excel. In this example, there is no clear winner in terms of both speed and accuracy simultaneously. If speed is prioritized, standard PINN (L-BFGS) is best; if accuracy is prioritized, boosted PINN (Adam + Newton) is best. 

\begin{figure}[ht]
    \centering
    \begin{subfigure}[t]{0.48\textwidth}
        \centering
        \includegraphics[width=\textwidth]{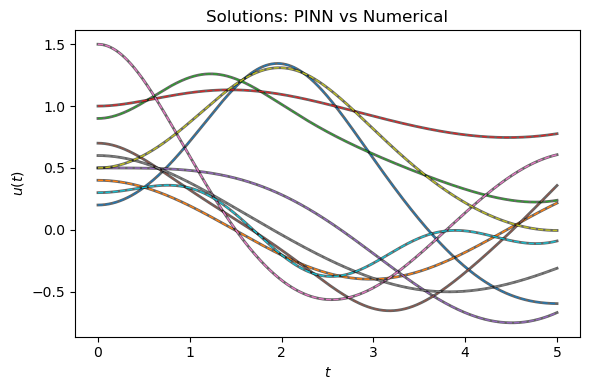}
        \caption{Boosted PINN (Adam + Newton) vs. numerical solution.}
        \label{fig:duffing_solution}
    \end{subfigure}
    \hfill
    \begin{subfigure}[t]{0.48\textwidth}
        \centering
        \includegraphics[width=\textwidth]{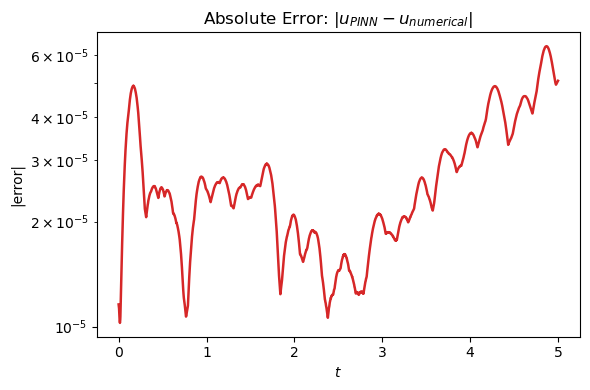}
        \caption{Error between Boosted PINN and numerical solution.}
        \label{fig:duffing_solution_error}
    \end{subfigure}
    \caption{Boosted PINN (Adam + Newton) results for the Duffing ODE.}
    \label{fig:duffing_bundle}
\end{figure}

The left-hand plot in Figure \ref{fig:duffing_bundle} shows the PINN's solution for the ODE, which converges to the numerical solution. The right-hand plot is the absolute error, defined simply as the absolute difference between the PINN solution and the numerical solution, $|\hat{u}_{PINN} - \hat{u}_{numerical}|$. Most of the error occurs later in the time domain. Since this is an IVP, it makes sense that the PINN has the most trouble learning dynamics further away from the initial conditions.

\begin{figure}[ht]
    \centering
    \includegraphics[width=1.0\textwidth]{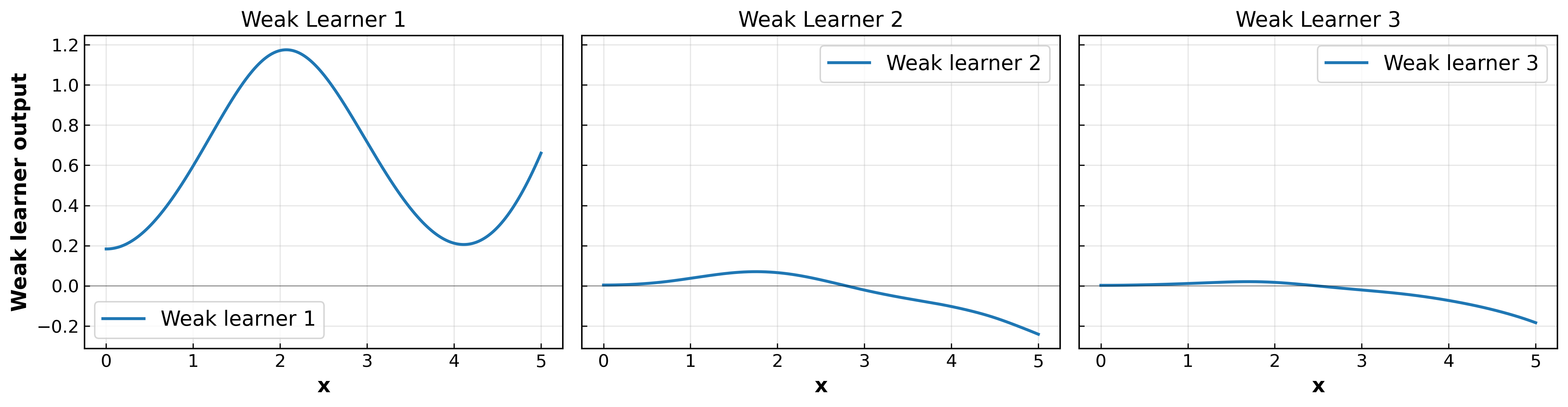}
    \caption{Boosted PINN (Adam) weak learners.}
    \label{fig:duffing_weak_learner}
\end{figure}

Figure \ref{fig:duffing_weak_learner} illustrates individual weak learners $h_k$ for $k = 1, 2, 3$ obtained by solving~\eqref{eq:restricted_problem}. The weak learners are combined additively with the corresponding stage weight $\alpha_k$. The final boosted PINN solution is the linear combination of all the weak learners. Only the first three weak learners are shown for illustrative purposes (out of 20 total stages), since $h_k \to 0$ as the stages progress and the ODE residual approaches zero.

\subsubsection{Nonlinear Reaction–Diffusion Equation}
\label{subsubsection:nrd_equation}
Next, we consider stiff and nonstiff NRD equations, both posed as BVPs. The NRD equation is
\begin{equation}
    -u''(x) + \kappa \bigl(u^3(x) - u(x)\bigr) = f(x), 
    \qquad x \in (0,1),
    \label{eq:stiff-reaction-bvp}
\end{equation}
with $f(x) \equiv 0$, subject to Dirichlet boundary conditions
$$
    u(0) = a = 0.0, 
    \qquad 
    u(1) = b = 2.0.
$$
Here, the second derivative term $-u''(x)$ represents diffusion with unit diffusivity, while the cubic nonlinearity $\kappa (u^3 - u)$ models a stiff reaction term. The parameter $\kappa \gg 1$ controls the strength of the nonlinearity and induces stiffness, leading to sharp spatial transitions and multiple scales in the solution. This boundary value problem serves as a challenging benchmark for PINN-based approaches due to its nonlinearity and stiffness. The purpose of this experiment is to show boosted PINN performance on nonlinear BVPs for both stiff and nonstiff regimes. 

We generate the numerical reference solution by reducing the second-order BVP to a first-order ODE system and solving it with \lstinline[language=Python]{solve_bvp()} at a tolerance of $1.0 \times 10^{-8}$. To aid convergence in the stiff regime, we initialize the solver with a tanh-based guess that smoothly interpolates between the boundary values and whose length scale reflects the expected sharp interior transition.


\paragraph{Results, \boldmath{$\kappa = 10$}} For the nonstiff case, we set $\kappa = 10$ in~\eqref{eq:stiff-reaction-bvp}. Comparing the performance of the standard PINNs to the boosted PINN, we see the standard PINNs never achieve values below $1.0 \times 10^{-2}$ on any of the metrics. The best-performing model with respect to MSE and relative $L^2$ error is the boosted PINN with conjugate gradients, at $1.32 \times 10^{-3}$ and $1.30 \times 10^{-3}$, respectively. The boosted PINN with Newton has the lowest residual norm, with a value of $7.36 \times 10^{-2}$. The results are somewhat surprising given that the equation is not stiff. What we are observing is the ability of the boosted PINN to achieve high performance with a very small architecture.

As for convergence, the boosted PINN using Adam alone has the fastest convergence time at 0.23 seconds. The boosted PINN with Adam + CG converges in fewer epochs (434), but its wall-clock time is slower than Adam alone, due to the extra computational cost of computing conjugate gradients. It nonetheless offers a performance benefit by requiring fewer epochs to converge.

\begin{table}[ht]
    \centering
    \begin{threeparttable}
    \begin{tabular}{lcc}
        \toprule
        & \multicolumn{2}{c}{Monolithic PINN} \\
        \cmidrule(lr){2-3}
        Metric & Adam & L-BFGS \\
        \midrule
        Residual Norm        & $5.05 \times 10^{-1}$ & $4.69 \times 10^{-2}$ \\
        MSE                  & $7.66 \times 10^{-2}$ & $1.03 \times 10^{-1}$ \\
        Relative $L^2$ Error    & $7.53 \times 10^{-2}$ & $1.03 \times 10^{-1}$ \\
        Training Time (s)    & -                      & -                      \\
        Number of Iterations & -                      & -                      \\
        Learning Rate        & $5.00 \times 10^{-4}$ & $1.0$                 \\
        Collocation Points   & $2{,}000$             & $2{,}000$             \\
        Network Size         & $593$                 & $593$                 \\
        \bottomrule
    \end{tabular}
    \caption{Standard PINN performance on the NRD equation.}
    \label{tab:nrd_standard}
    \begin{tablenotes}
        \scriptsize
        \item \text{--} indicates that the standard PINN did not converge; iteration count and training time are therefore not reported.
    \end{tablenotes}
    \end{threeparttable}
\end{table}

\begin{table}[ht]
    \centering
    \begin{tabular}{lccc}
        \toprule
        & \multicolumn{3}{c}{Boosted PINN} \\
        \cmidrule(lr){2-4}
        Metric & Adam & Adam + CG & Adam + Newton \\
        \midrule
        Residual Norm        & $1.90 \times 10^{-1}$ & $2.35 \times 10^{-1}$ & $7.36 \times 10^{-2}$ \\
        MSE                  & $1.45 \times 10^{-3}$ & $1.32 \times 10^{-3}$ & $2.14 \times 10^{-3}$ \\
        Relative $L^2$ Error    & $1.42 \times 10^{-3}$ & $1.30 \times 10^{-3}$ & $2.10 \times 10^{-3}$ \\
        Training Time (s)    & $0.23$                & $1.61$                & $40.28$               \\
        Number of Iterations & $658$                 & $434$                 & $462$                 \\
        Learning Rate        & $10^{-2}$             & $10^{-2}$             & $10^{-2}$             \\
        Collocation Points   & $2{,}000$             & $2{,}000$             & $2{,}000$             \\
        Network Size         & $593$                 & $593$                 & $593$                 \\
        \bottomrule
    \end{tabular}
    \caption{Boosted PINN performance on the NRD equation.}
    \label{tab:nrd_boosted}
\end{table}

\begin{figure}[ht]
    \centering
    \includegraphics[width=1.0\textwidth]{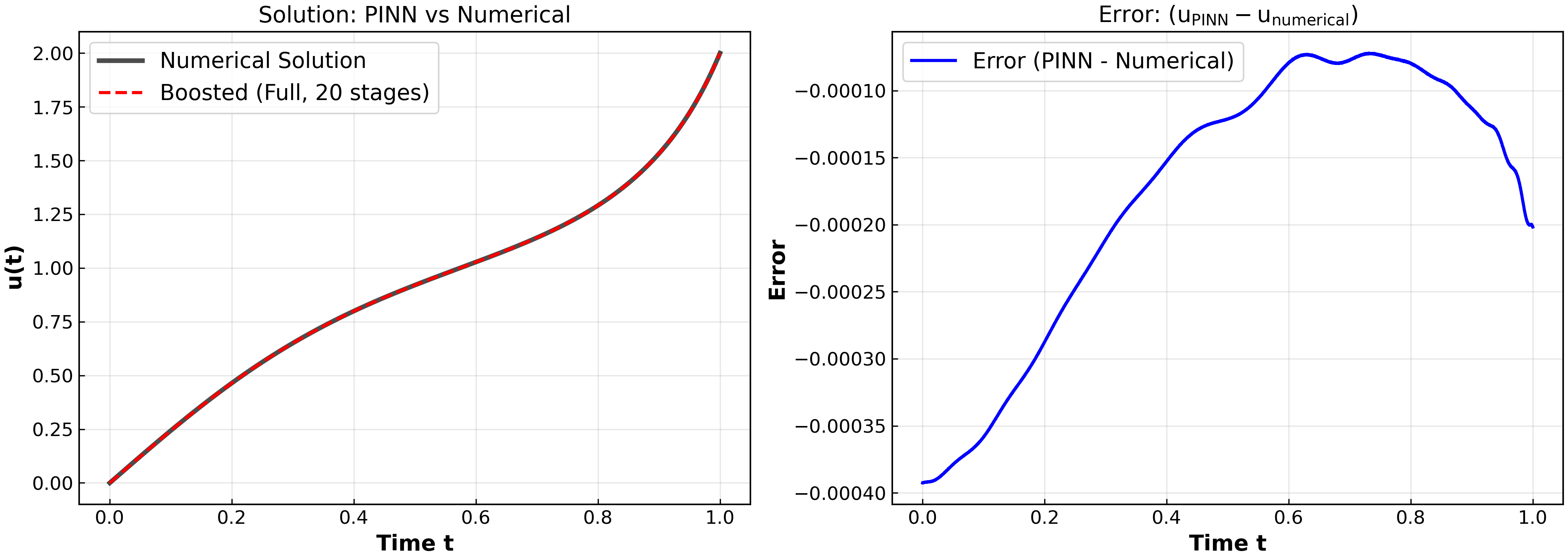}
    \caption{Boosted PINN (Adam + CG) solution vs.\ numerical solution.}
    \label{fig:nrd_solution_fit}
\end{figure}

\begin{figure}[H]
    \centering
    \includegraphics[width=1.0\textwidth]{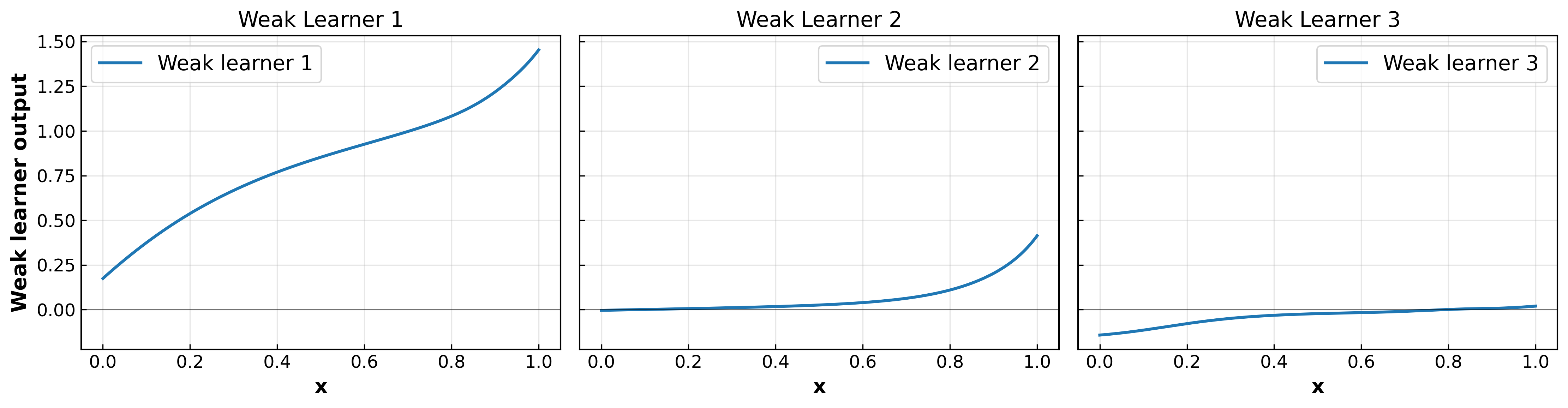}
    \caption{Boosted PINN weak learners for the nonstiff NRD equation.}
    \label{fig:nrd_weak_learners}
\end{figure}

Since the example is a BVP, the PINN's error mainly occurs in the interior of the domain, as shown in Figure~\ref{fig:nrd_solution_fit}. The loss function includes a boundary condition loss, which guides the PINN to the correct boundary values $u(0) = 0.0$ and $u(1) = 2.0$. The error is therefore expected to be lower at the boundaries. Figure~\ref{fig:nrd_weak_learners} shows a subset of the weak learners across the 20 stages. The same pattern emerges as in the Duffing example: as the ODE residual is reduced, $h_k \to 0$.

\paragraph{Results, \boldmath{$\kappa = 100$}} For the stiff case, we set $\kappa = 100$ in \eqref{eq:stiff-reaction-bvp}. In this regime, both standard PINNs fail to converge to the reference numerical solution; we therefore omit the iteration and training time metrics for both in Table~\ref{tab:nrd_stiff_results}. The stiffness of the equation leads to sharp spatial transitions and a highly ill-conditioned optimization landscape, making gradient-based training particularly challenging for the baseline model. However, the boosted PINN successfully converges in 0.27 seconds with 858 epochs. It achieved a residual norm of $2.26\times 10^{0}$, an MSE of $8.01\times 10^{-3}$, and a relative $L^2$ error of $8.01\times 10^{-3}$. Note that these values are identical because they are rounded. At four decimal places, the values are different. The relatively high residual norm, despite the model's low MSE and relative $L^2$ error, reflects the sensitivity of the residual to small remaining errors in the sharp transition region. Minor deviations near this region are amplified in the residual while contributing only marginally to the domain-averaged MSE.

\begin{table}[ht]
    \centering
    \begin{threeparttable}
    \begin{tabular}{lccc}
        \toprule
        & \multicolumn{2}{c}{Monolithic PINN} & Boosted PINN \\
        \cmidrule(lr){2-3}
        Metric & Adam & L-BFGS & Adam \\
        \midrule
        Residual Norm        & $3.53 \times 10^{0}$  & $4.08 \times 10^{-1}$ & $2.62 \times 10^{0}$  \\
        MSE                  & $1.00 \times 10^{0}$  & $9.27 \times 10^{-1}$ & $8.01 \times 10^{-3}$ \\
        Relative $L^2$ Error    & $1.00 \times 10^{0}$  & $9.27 \times 10^{-1}$ & $8.01 \times 10^{-3}$ \\
        Training Time (s)    & --                    & --                    & $0.27$                \\
        Number of Iterations & --                    & --                    & $858$                 \\
        Learning Rate        & $5.00 \times 10^{-4}$ & $1.0$                 & $1.00 \times 10^{-2}$ \\
        Collocation Points   & $2{,}000$             & $2{,}000$             & $2{,}000$             \\
        Network Size         & $593$                 & $593$                 & $593$                 \\
        \bottomrule
    \end{tabular}
    \caption{Performance of standard and boosted PINN on the NRD equation 
    ($\kappa = 100$).}
    \label{tab:nrd_stiff_results}
    \begin{tablenotes}
        \scriptsize
        \item \text{--} indicates that the standard PINN did not converge; iteration count and training time are therefore not reported. 
    \end{tablenotes}
    \end{threeparttable}
\end{table}

\begin{figure}[ht]
    \centering
    \includegraphics[width=1.0\textwidth]{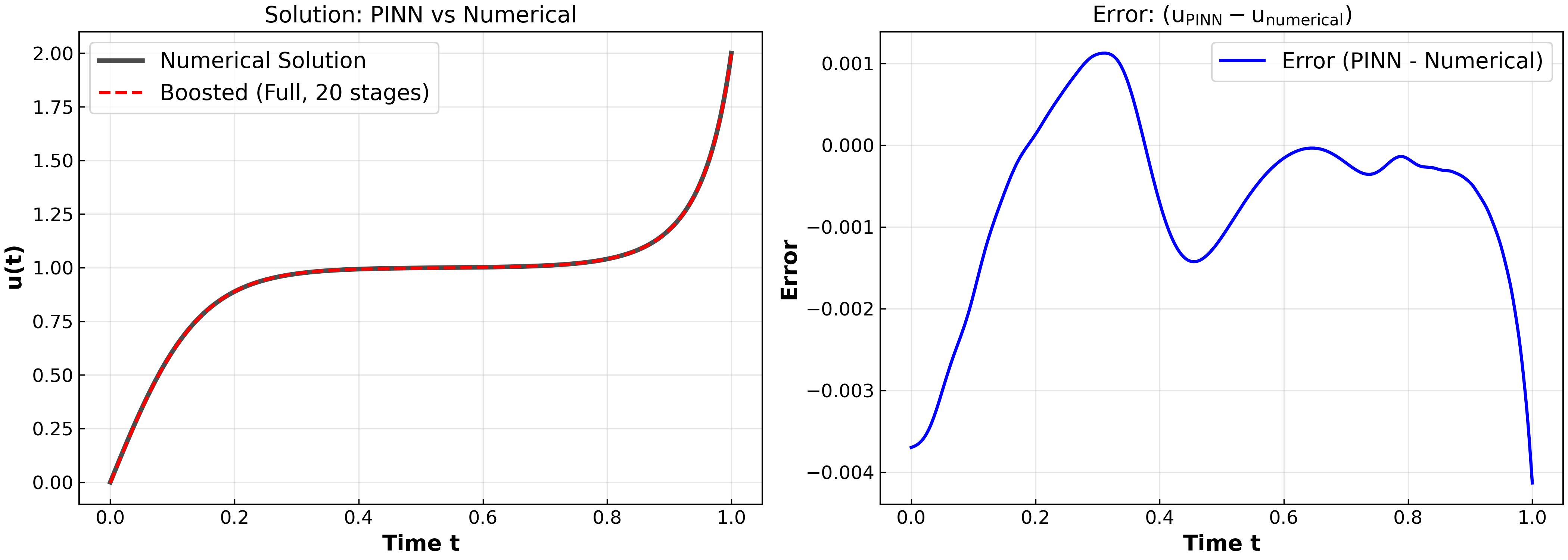}
    \caption{Boosted PINN (Adam) solution vs. numerical solution}
    \label{fig:nrd_stiff_solution_fit}
\end{figure}

Figure~\ref{fig:nrd_stiff_solution_fit} shows the boosted PINN solution with the Adam optimizer, along with the corresponding error plot. The solution closely follows the numerical solution. Most of the error occurs in the interior of the domain, as shown by the right-hand plot. The rationale is the same as with the nonstiff case. For a BVP, the PINN has a corresponding loss for the boundary conditions, which allows it to more easily determine $u(0) = 0.0$ and $u(1) = 2.0$.

\begin{figure}[ht]
    \centering
    \includegraphics[width=1.0\textwidth]{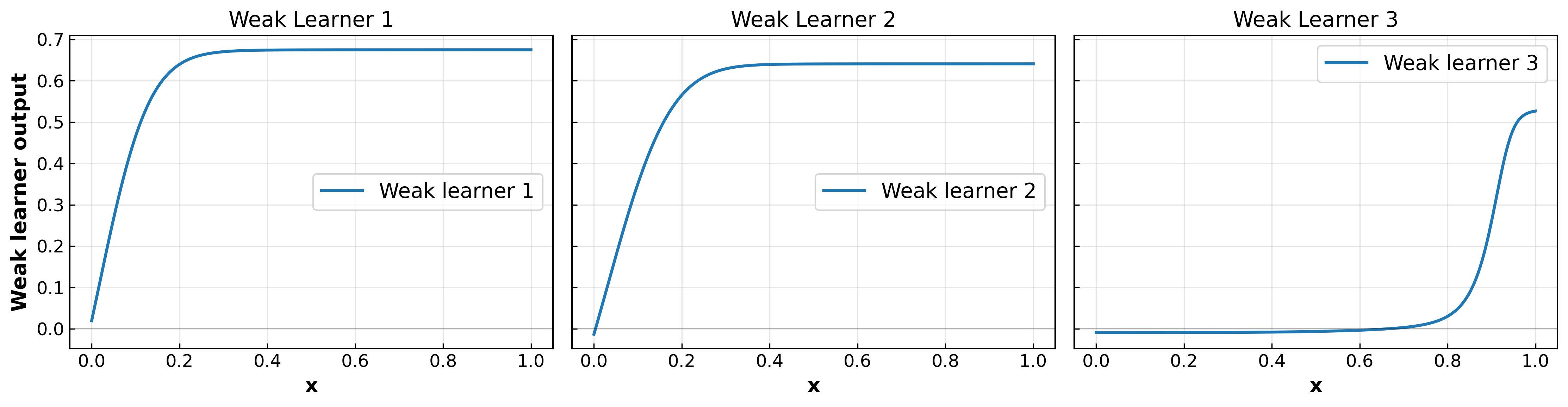}
    \caption{Boosted PINN weak learners}
    \label{fig:nrd_stiff_weak_learners}
\end{figure}

The weak learners are shown in Figure~\ref{fig:nrd_stiff_weak_learners}. Unlike the nonstiff case, there is no rapid decay towards zero because the problem is stiff, making it more difficult to solve. It takes more weak learners for the boosted PINN to converge to the solution: 858 epochs for the stiff problem versus 658 for the nonstiff problem.

\subsubsection{Van der Pol Equation}
\label{subsec:van_der_pol}

The final ODE example is the Van der Pol equation, posed as an IVP. The differential equation is given by
\begin{equation}
    \frac{d^2u}{dt^2} - \mu(1-u^2)\frac{du}{dt} + u = 0, 
    \qquad u(0) = 2, \quad 
    \left.\frac{du}{dt}\right|_{t=0} = 0.
\end{equation}
The equation is solved over the domain $t \in [0, 8]$. The parameter $\mu$ controls the stiffness of the equation; the larger $\mu$, the stiffer the problem. The equation is presented here in dimensional form; we initially attempted to fit the boosted PINN on the nondimensionalized form, using transfer learning from one boosted ensemble to another trained at a slightly higher value of $\mu$, but this approach did not work, since the stiff region of the domain shifts as $\mu$ increases, making transfer learning on the correction stages ineffective.

Note that the standard PINN trained with both the Adam and L-BFGS optimizers fails to converge for this problem and is therefore omitted from the results. For highly stiff problems such as Van der Pol with large $\mu$, where even the boosted PINN falls short, the recommended approach remains the curriculum strategy of~\citep{seiler2025stiff}, which uses transfer learning across incrementally increasing $\mu$ values.

The architecture of the PINN for this example uses $\sin$ activations with a hyperparameter $\omega$ for the frequency of the $\sin$ function. A learning rate scheduler is used, decaying from $10^{-2}$ to the final value shown in Table~\ref{tab:vdp_boosted}. When training the boosted PINN on the stiff Van der Pol equation, stage 0 is extremely weak and fails to converge to any meaningful fit. This causes training to become unstable, leaving the correction stages unable to recover the correct solution. To address this, we use retry logic with frequency annealing. The retry logic reruns the stage 0 fit five times and selects the fit with the lowest residual norm. Frequency annealing starts $\omega$ at a value near $1$ for the first epoch and gradually increases it to $30$ at the final epoch.

\begin{table}[h]
\centering
\begin{threeparttable}
\begin{tabular}{lccc}
    \toprule
    Metric
        & $\mu = 2$
        & $\mu = 3$
        & $\mu = 4$ \\
    \midrule
    Optimizer
        & Adam & Adam & Adam \\
    Residual Norm
        & $9.26\times10^{-3}$
        & $1.77\times10^{-2}$
        & $1.453$ \\
    MSE
        & $1.36\times10^{-3}$
        & $5.71\times10^{-3}$
        & $2.94\times10^{-1}$ \\
    Relative $L^2$ Error
        & $9.05\times10^{-4}$
        & $3.76\times10^{-3}$
        & $1.81\times10^{-1}$ \\
    Training Time (s)
        & $2{,}736.71$
        & $9{,}474.38$
        & -- \\
    Number of Iterations
        & $66{,}296$
        & $134{,}234$
        & $98{,}031$ \\
    Learning Rate
        & $10^{-2} \to 1.00\times10^{-4}$
        & $10^{-2} \to 2.5\times10^{-5}$
        & $10^{-2} \to 1.25\times10^{-5}$ \\
    Collocation Points
        & $3{,}000$ & $3{,}000$ & $3{,}000$ \\
    Network Size (parameters)
        & $49{,}921$ & $49{,}921$ & $49{,}921$ \\
    \bottomrule
\end{tabular}
\caption{Boosted PINN performance on the Van der Pol Equation.}
\label{tab:vdp_boosted}
\begin{tablenotes}
    \scriptsize
    \item \text{--} indicates that the PINN did not converge; iteration count and training time are therefore not reported.
\end{tablenotes}
\end{threeparttable}
\end{table}

The results show convergence for $\mu = 2$ and $\mu = 3$, with MSE $< 10^{-2}$ in both cases. Convergence takes much longer for this example than for any example in this paper. For $\mu=2.0$, convergence takes 2,736.71 seconds. For $\mu=3.0$, convergence takes 9,474.38 seconds. This example is challenging for two reasons. First, it is an initial value problem, so the PINN must find the solution without knowing the end state; any deviation early in the solution trajectory can lead to an inaccurate solution. 

At $\mu = 4$, training becomes unstable, reflected in the degradation of the residual norm and MSE. Over independent runs, the MSE for $\mu = 4$ ranged from $2.37 \times 10^{-4}$ to $1.09$, indicating that the boosted PINN converges for some runs but not others. The MSE and relative $L^2$ error were computed against a reference solution from the Radau solver, implemented using \lstinline[language=Python]{solve_ivp()}, with an absolute tolerance of $1.0 \times 10^{-6}$ and a relative tolerance of $1.0 \times 10^{-3}$.

\begin{figure}[h]
    \centering
    \begin{subfigure}[b]{0.48\textwidth}
        \centering
        \includegraphics[width=\textwidth]{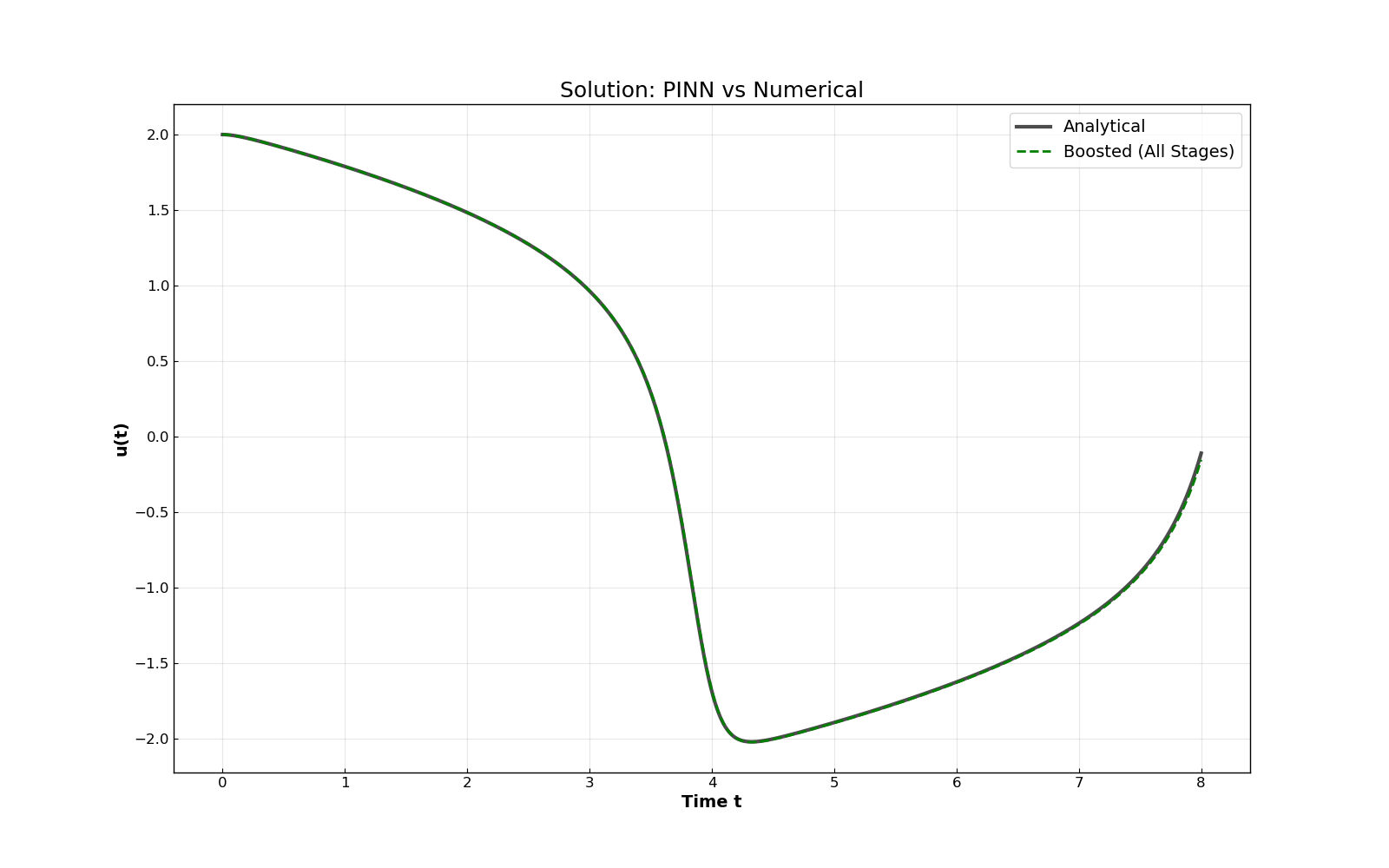}
        \caption{Boosted PINN solution vs.\ numerical solution.}
        \label{fig:vdp_best_fit}
    \end{subfigure}
    \hfill
    \begin{subfigure}[b]{0.48\textwidth}
        \centering
        \includegraphics[width=\textwidth]{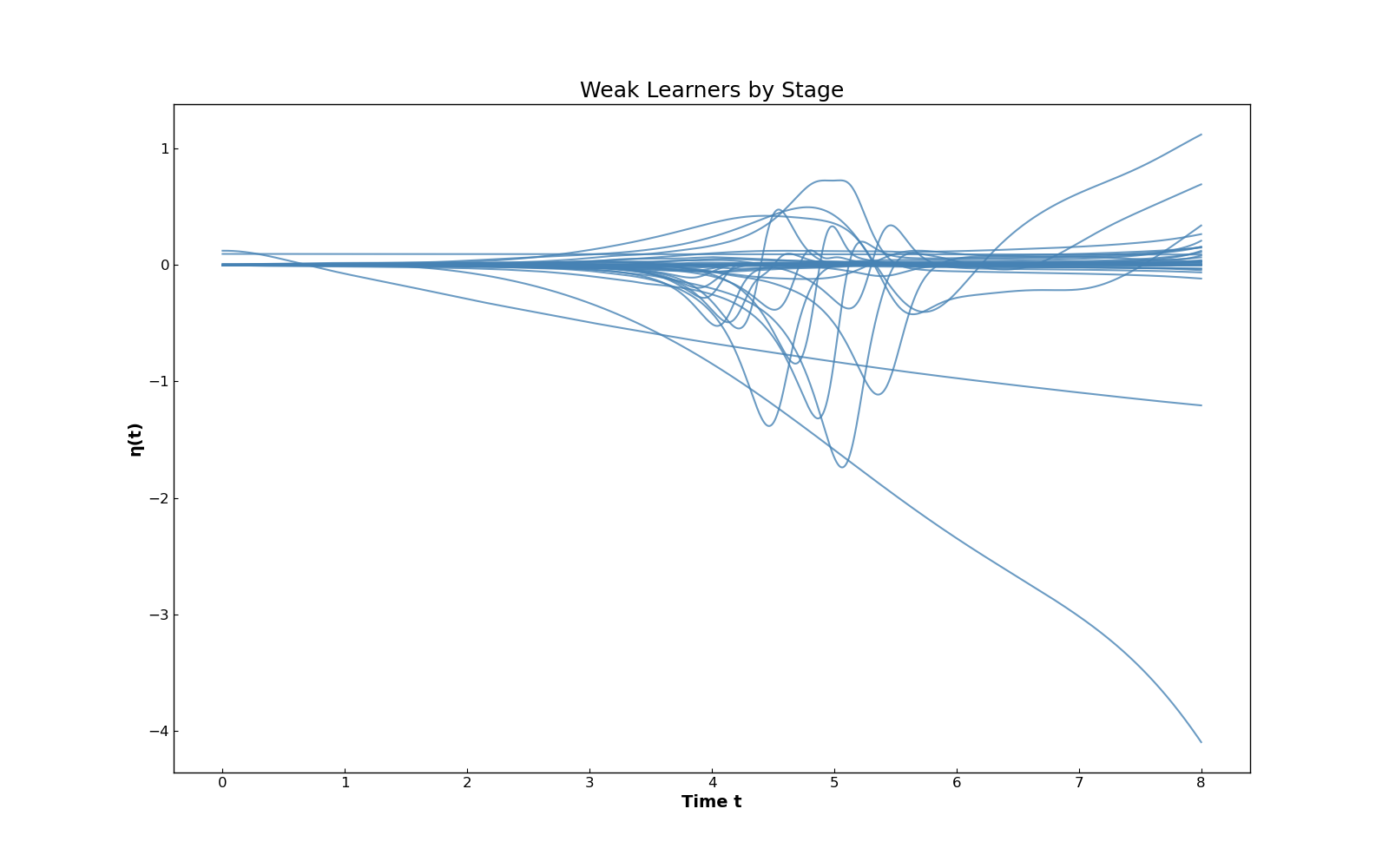}
        \caption{Boosted PINN weak learners.}
        \label{fig:vdp_weak_learners}
    \end{subfigure}
    \caption{Van der Pol oscillator ($\mu$ = 3.0): boosted PINN solution and weak learners.}
    \label{fig:vdp_boosted_pinn}
\end{figure}

Figure~\ref{fig:vdp_boosted_pinn} shows the boosted PINN solution and the weak learners across 40 stages. The solution closely tracks the numerical reference solution, demonstrating convergence. The weak learners in the right-hand plot show that most of the corrections are concentrated on the stiff region of the domain. This is the most challenging example in this paper for the boosted PINN. Despite this, the boosted PINN converges on the dimensional Van der Pol equation for $\mu = 3$, without transfer learning from models trained on smaller $\mu$ values and without relying on nondimensionalization. To our knowledge, no prior PINN-based method has demonstrated convergence on the dimensional Van der Pol equation at $\mu = 3$ under these conditions. Prior work has shown failure to converge on this problem~\citep{zhai-2021}, while other approaches have succeeded only by decomposing the equation into a system of ODEs with nondimensionalization~\citep{tarancon-alvarez-2025}.

\subsection{System of ODEs}

In this section, we move away from scalar ODEs and demonstrate the performance of the boosted PINN on a nonlinear coupled IVP, the Lotka--Volterra predator--prey model. It is a 2D autonomous system exhibiting closed periodic orbits in the phase plane. The system is characterized by the following equations:
\begin{equation}
    \begin{cases}
    \dfrac{dx}{dt} = \alpha x - \beta x y, \qquad x(0) = x_0 = 10.0, \\[10pt]
    \dfrac{dy}{dt} = -\gamma y + \delta x y, \qquad y(0) = y_0 = 5.0,
    \end{cases}
    \label{eq:lotka-volterra}
\end{equation}
where $x(t)$ denotes the prey population and $y(t)$ denotes the predator population. The parameter $\alpha > 0$ is the natural growth rate of the prey in the absence of predators, and $\beta > 0$ is the rate at which predators destroy prey, governed by the nonlinear interaction term $xy$. For the predator equation, $\gamma > 0$ is the natural death rate of predators in the absence of prey, and $\delta > 0$ is the rate at which predators increase by consuming prey. The nonlinearity of the system arises entirely from the bilinear coupling terms $\beta xy$ and $\delta xy$, which couple the two equations and prevent them from being solved independently. 

The parameter values are set as follows: $\alpha = 1.0$, $\beta = 0.1$, $\gamma = 1.5$, and $\delta = 0.075$. The non-trivial equilibrium of the system is located at $(x^*, y^*) = (\gamma/\delta,\, \alpha/\beta) = (20, 10)$, and all trajectories initialized away from this point form closed orbits, making it a challenging benchmark for PINNs due to the oscillatory nature of the solution over long time horizons. It is known that PINNs struggle with long time horizons~\cite{wang2022respecting}; as such, we restrict the time horizon to approximately one cycle, with $t \in [0.0, 5.5]$.

Since it is a 2D system, the output dimension of the PINN will now be two. For the numerical reference solution, we used a Runge--Kutta 4(5) solver, with a relative tolerance of $1.0 \times 10^{-10}$ and an absolute tolerance of $1.0 \times 10^{-12}$, using \lstinline[language=Python]{solve_ivp()}.

\begin{table}[h]
    \centering
    \begin{threeparttable}
    \begin{tabular}{lccc}
        \toprule
        & \multicolumn{2}{c}{Monolithic PINN} & \\
        \cmidrule(lr){2-3}
        Metric & Adam & L-BFGS & Boosted PINN \\
        \midrule
        Residual Norm          & $1.21 \times 10^{-2}$ & $2.50 \times 10^{-1}$ & $5.62 \times 10^{-3}$ \\
        MSE                    & $1.74 \times 10^{-2}$ & $1.21 \times 10^{0}$  & $2.92 \times 10^{-3}$ \\
        Relative $L^2$ Error      & $9.52 \times 10^{-4}$ & $6.62 \times 10^{-2}$ & $1.60 \times 10^{-4}$ \\
        Training Time (s)      & --                    & --                    & $38.06$               \\
        Iterations             & --                    & --                    & $26{,}400$            \\
        Learning Rate          & $10^{-3}$             & $1.0$                 & $10^{-3}$             \\
        Collocation Points     & $1{,}000$             & $1{,}000$             & $1{,}000$             \\
        Network Parameters     & $3{,}298$             & $3{,}298$             & $3{,}298$             \\
        \bottomrule
    \end{tabular}
    \caption{Performance comparison on the Lotka--Volterra predator--prey system.}
    \label{tab:lv_comparison}
    \begin{tablenotes}
        \scriptsize
        \item \text{--} indicates that the standard PINN did not converge; iteration count and training time are therefore not reported. 
    \end{tablenotes}
    \end{threeparttable}
\end{table}


Neither of the standard PINNs converges under our fixed MSE threshold. Notably, the standard PINN with Adam achieves a low relative $L^2$ error despite its MSE remaining above the convergence criterion. This discrepancy arises because relative $L^2$ error is normalized by the magnitude of the reference solution, while MSE is not. Since the Lotka--Volterra populations range over tens of units, even a modest absolute error is small in relative terms but can still exceed the unnormalized MSE threshold, which was calibrated for problems with solutions ranging roughly between -1 and 1. This suggests that relative $L^2$ error is the more reliable metric for judging model performance on this example. However, MSE remains useful as an absolute measure of error, which may matter more in settings where the physical magnitude of the error is of primary interest. Overall, the boosted PINN is best across all metrics, achieving a residual norm of $5.62 \times 10^{-3}$, an MSE of $2.92 \times 10^{-3}$, and a relative $L^2$ error of $1.60 \times 10^{-4}$. The boosted PINN  converges, on average, in 38.06 seconds and 26,400 epochs.

\begin{figure}[H]
    \centering
    \includegraphics[width=1.0\textwidth]{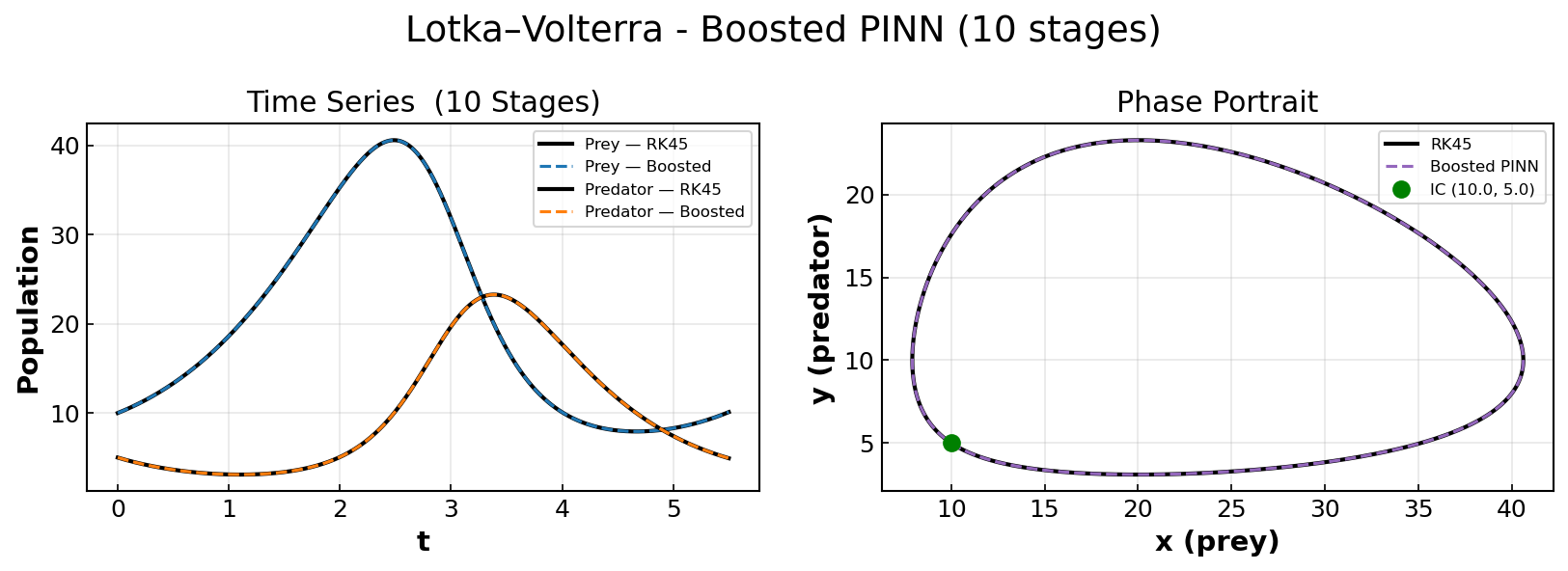}
    \caption{Boosted PINN solution vs. numerical solution}
    \label{fig:lotka_volterra_solution}
\end{figure}

\begin{figure}[H]
    \centering
    \includegraphics[width=1.0\textwidth]{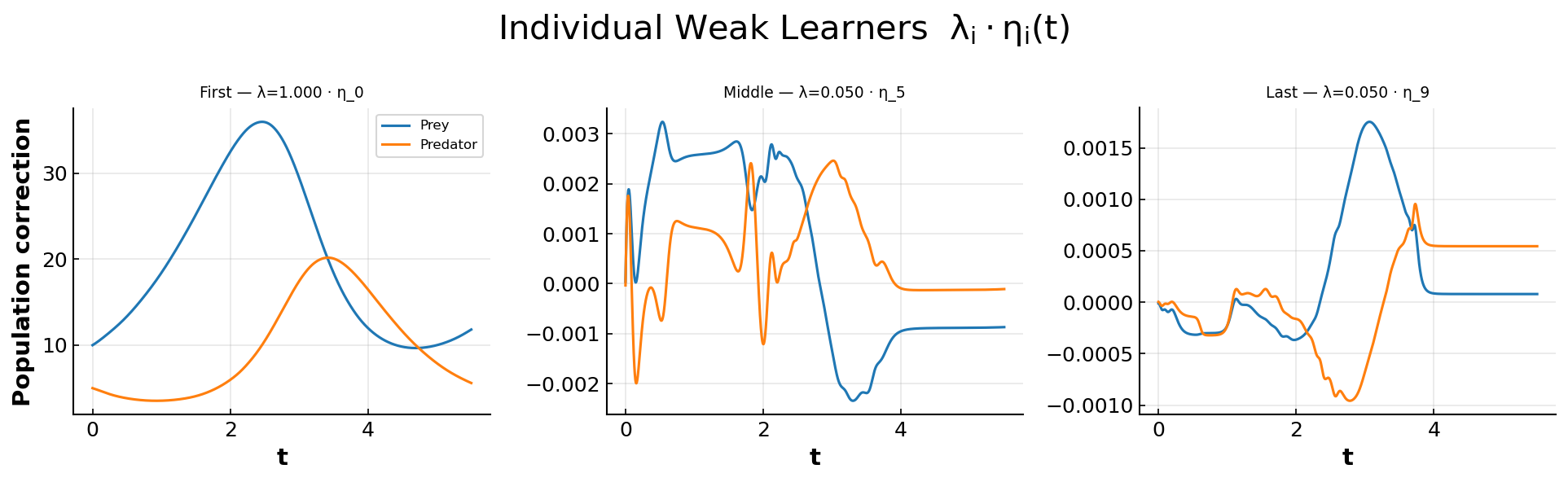}
    \caption{Boosted PINN weak learners}
    \label{fig:lotka_volterra_weak_learners}
\end{figure}

Figure~\ref{fig:lotka_volterra_solution} shows the boosted PINN solution against the numerical reference solution, as both a time trajectory and a phase portrait. The boosted PINN solution closely aligns with the numerical solution. Figure~\ref{fig:lotka_volterra_weak_learners} shows the weak learners for stage 0, stage 6, and stage 10. The pattern we observe is that stage 0 captures the overall solution, but with noticeable error. With each new correction, the solution shifts toward the true solution. As the physics residual decreases, the correction magnitude $h_k \to 0$, consistent with the decreasing order of magnitude observed between stages. Stage 6 makes adjustments on the order of $1.0 \times 10^{-3}$, while stage 10's adjustments are on the order of $1.0 \times 10^{-4}$. This is exactly the behavior described in Section~\ref{subsection:boosted_pinn_framework} and observed in prior examples.
 
\subsection{PDE Examples}
Having validated the boosted PINN on ODE examples, we now extend to PDE examples to demonstrate broader applicability. The two PDE examples are Burgers' equation and the Allen-Cahn equation. Each equation has a parameter that controls the level of stiffness, and we show results for both stiff and nonstiff regimes. For this section, every result shown in the tables is averaged over a minimum of 10 independent seeds. Similar to the ODE examples, the number of epochs for both the standard PINNs and the boosted PINN is fixed and equal, meaning the total epochs across all stages equals the number of epochs the standard PINN is trained on. If a PINN does not achieve an MSE less than $1.0 \times 10^{-2}$, within the given epoch budget, we say it does not converge. The interpretation of the convergence metrics (training time and iterations) remains the same as with the ODE examples. The residual norm, MSE, and relative $L^2$ error reported are the best values achieved by each respective model. In all PDE examples, we set the batch size equal to the training set. In other words, there is no mini-batch gradient descent. For parameter parity, each weak learner has the same architecture as the standard PINN.

\begin{table}[ht]
    \centering
    \begin{tabular}{llcc}
        \toprule
        Type & Name & Parameter Regime & Loss Linearized \\
        \midrule
        Nonlinear                              & Burgers 1D          & $\nu = 1.0$            & No  \\
        Stiff Nonlinear                        & Inviscid Burgers 1D & $\nu = 0.002$       & No  \\
        Nonlinear Reaction Diffusion            & Allen-Cahn          & $D = 1.0$              & Yes \\
        Stiff Nonlinear Reaction Diffusion      & Allen-Cahn          & $D = 10^{-4}$          & Yes \\
        \bottomrule
    \end{tabular}
    \caption{Overview of PDE examples.}
    \label{tab:pde_examples}
\end{table}

\paragraph{Transfer Learning for PDE Examples}
As with the ODE examples, the weak learner networks at each boosting stage transfer weights from the previous stage's network, with the final two layers re-initialized using Xavier initialization~\citep{pmlr-v9-glorot10a} and rescaled by a stage-specific scaling factor $\xi_k$, as described in Section~\ref{subsection:boosted_pinn_framework}. Table~\ref{tab:pde_transfer_learning} provides the scaling factor used for each problem and regime. The ablation results in Section~\ref{sec:ablation_study} demonstrate that transfer learning improves boosted PINN performance.

\begin{table}[H]
    \centering
    \begin{tabular}{llll}
        \toprule
        Type & PDE Name & Regime & 
        Scale Factor ($\xi$) \\
        \midrule
        Nonlinear & Burgers' 1D & high $\nu$ & 0.01 \\
        Stiff Nonlinear & Burgers' 1D & low $\nu$ & 0.01 \\
        Nonlinear Reaction-Diffusion & Allen-Cahn (Dirichlet) & large $D$ & 0.5 \\
        Stiff Nonlinear Reaction-Diffusion & Allen-Cahn (Periodic) & small $D$ & 0.5 \\
        \bottomrule
    \end{tabular}
    \caption{Scale factors used for transfer learning across PDE examples.}
    \label{tab:pde_transfer_learning}
\end{table}

\subsubsection{Burgers' Equation}

The Burgers' equation is given by
\begin{equation}
    \frac{\partial u}{\partial t} + u \frac{\partial u}{\partial x} 
    = \nu \frac{\partial^2 u}{\partial x^2},  
    \quad x \in [-1,1],
    \label{eq:burgers}
\end{equation}
where $u(x,t)$ is the velocity field, $\nu$ is the viscosity, $x$ is the spatial coordinate, and $t$ is time. The initial condition is
\begin{equation}
    u(x,0) = -\sin(\pi x).
\end{equation}
The boundary conditions are homogeneous Dirichlet boundary conditions
\begin{equation}
    u(-1,t) = u(1,t) = 0,  \qquad t \in [0, 0.5].
\end{equation}
The smaller the value of $\nu$, the more difficult the equation is to solve, due to the formation of sharp gradients (shock waves). 

To obtain the MSE and relative $L^2$ error, we require a numerical reference. We generate the reference numerical solution by discretizing space on a fine uniform grid, converting the PDE into a system of ODEs using 4th-order central finite differences in the interior (2nd-order near the boundaries). This system is then integrated in time with a high-accuracy Radau method (relative tolerance = $10^{-12}$, absolute tolerance = $10^{-14}$), producing dense-output solution values across many time points. 


\paragraph{Results, \boldmath{$\nu = 0.1$}}
We start with the nonstiff regime. The best model is the standard PINN with L-BFGS, outperforming the boosted PINN on MSE, relative $L^2$ error, and residual norm. However, the boosted PINN converges faster, reaching convergence in 22.03 seconds and 600 epochs, compared to 48.05 seconds and 8,009 function calls for the standard PINN with L-BFGS. For L-BFGS, we report the total number of closure calls until convergence to make the comparison with Adam accurate. These results are consistent with our earlier findings: for simple problems, L-BFGS achieves better accuracy, as we saw with the nonstiff NRD equation (Section~\ref{subsubsection:nrd_equation}), while the boosted PINN converges the quickest. The standard PINN with Adam does not converge.

\begin{table}[ht]
    \centering
    \begin{threeparttable}          
        \begin{tabular}{lccc}
            \toprule
            & Standard PINN & Standard PINN 
            & Boosted PINN \\
            \midrule
            Optimizer          & Adam                  & L-BFGS                
                               & Adam                  \\
            Residual Norm      & $3.74 \times 10^{-2}$ & $5.50 \times 10^{-3}$ 
                               & $8.69 \times 10^{-3}$ \\
            MSE                & $5.01 \times 10^{-1}$ & $1.82 \times 10^{-4}$ 
                               & $6.83 \times 10^{-4}$ \\
            Relative $L^2$ Error  & $9.07 \times 10^{-1}$ & $3.29 \times 10^{-4}$ 
                               & $1.24 \times 10^{-3}$ \\
            Training Time (s)  & --                    & $48.05$              
                               & $22.03$               \\
            Iterations         & --                    & $8{,}009$\tnote{*}   
                               & $600$                 \\
            Learning Rate      & $10^{-2}$             & $1.0$             
                               & $10^{-2}$             \\
            Collocation Points & $5{,}000$             & $5{,}000$             
                               & $5{,}000$             \\
            Network Parameters & $2{,}241$             & $2{,}241$            
                               & $2{,}241$             \\
            \bottomrule
        \end{tabular}
        \caption{Performance comparison on Burgers' equation 
        ($\nu = 0.1$).}
        \label{tab:burgers_performance}
        \begin{tablenotes}
        \scriptsize
        \item \text{--} indicates that the standard PINN did not converge; iteration count and training time are therefore not reported.
        \item[*] L-BFGS ran for $8{,}009$ function evaluations across $500$ iterations.
        \end{tablenotes}
    \end{threeparttable}
\end{table}

\begin{figure}[H]
    \centering
    \includegraphics[width=1.0\textwidth]{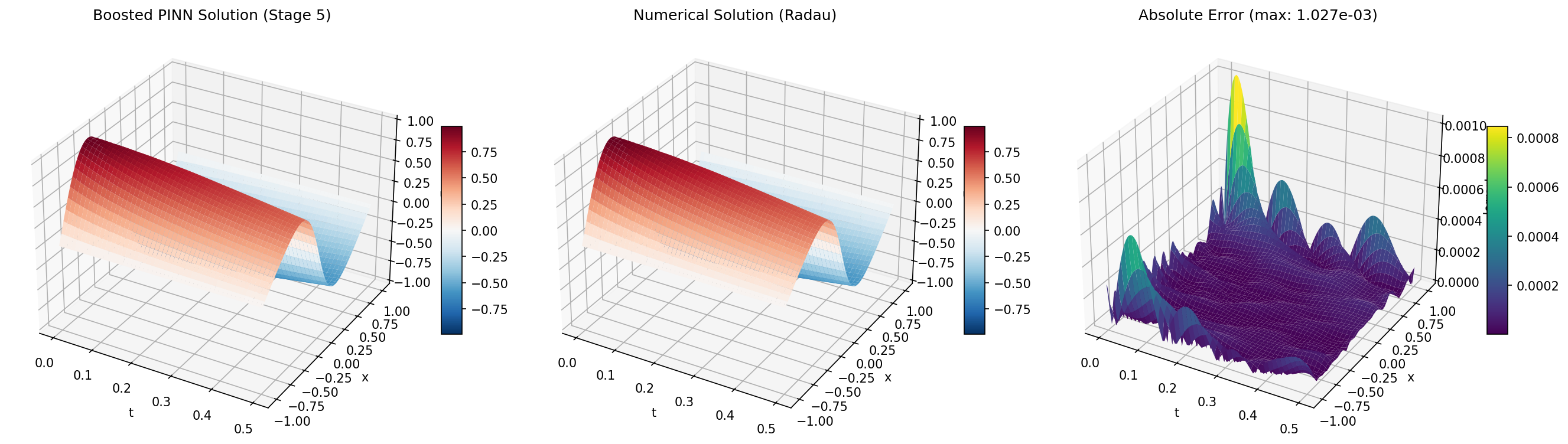}
    \caption{Boosted PINN solution vs.\ numerical solution for 
    Burgers' equation ($\nu = 0.1$).}
    \label{fig:burgers0.1_solution_fit}
\end{figure}

\begin{figure}[H]
    \centering
    \includegraphics[width=0.6\textwidth]{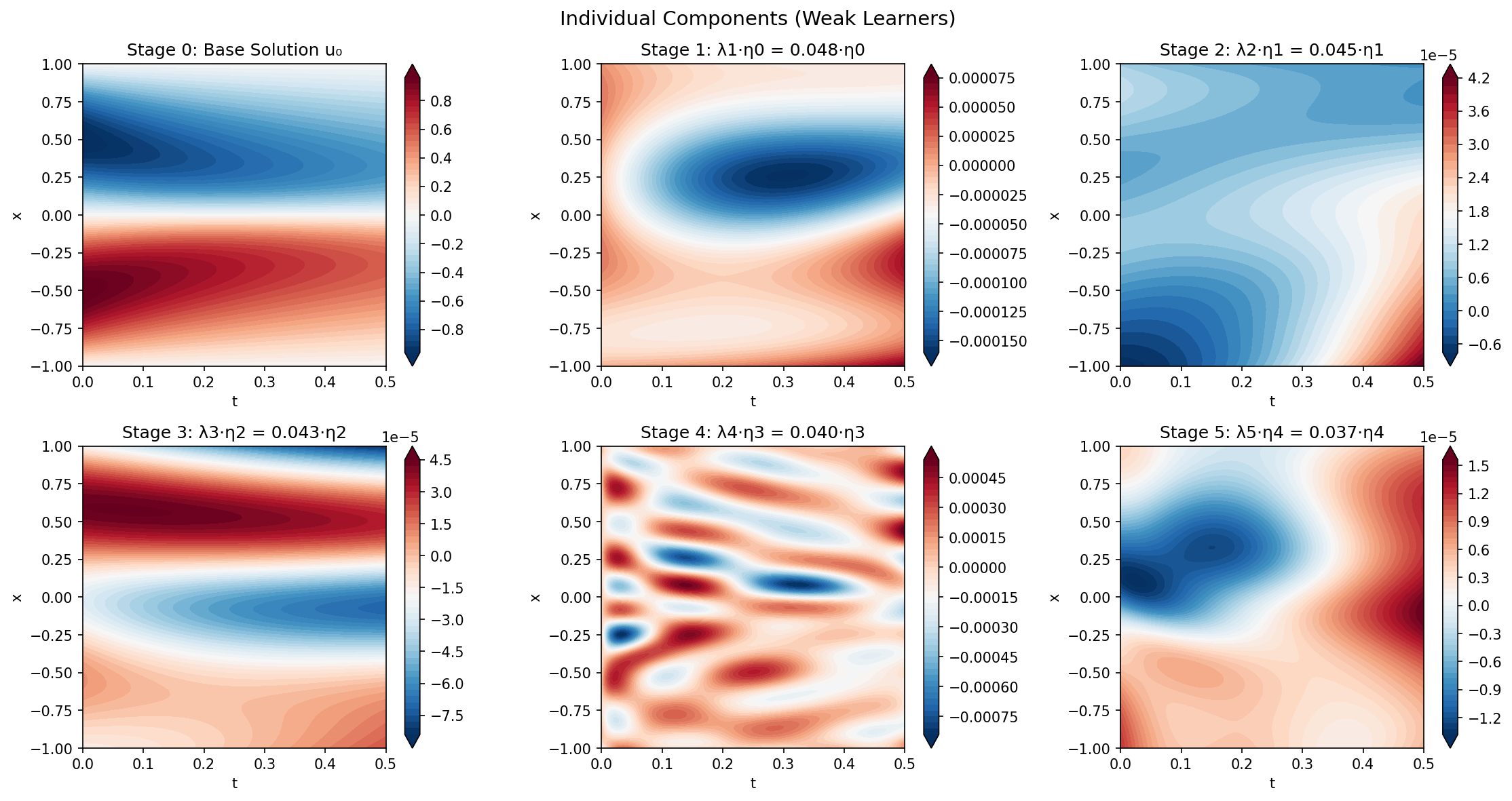}
    \caption{Boosted PINN weak learners for Burgers' equation 
    ($\nu = 0.1$).}
    \label{fig:burgers0.1_weak_learners}
\end{figure}

Figure~\ref{fig:burgers0.1_solution_fit} compares the boosted PINN solution to the numerical solution. Most of the error occurs at $t = 0$, suggesting a need for a larger weight on the initial condition loss. Figure~\ref{fig:burgers0.1_weak_learners} shows a subset of the weak learners. The corrections are of very small magnitude because stage 0 already performs well; the subsequent stages provide only marginal improvements to the residual norm and MSE, highlighting the simplicity of the example. Overall, there is no clear winner in this case; the preferred model depends on whether one prioritizes convergence speed or accuracy.

\paragraph{Results, \boldmath{$\nu = 0.002$}}
Next, we decrease $\nu$ to test the models under a stiff regime. In prior research, standard PINNs have been shown to converge with an MSE of order $10^{-4}$ for $\nu = \frac{0.01}{\pi} \approx 0.00318$ \cite{Raissi2019PINNs}. Note that our case is stiffer. Neither the standard PINN nor the boosted PINN converges to the criterion MSE $< 10^{-2}$; as a result, no training times or iteration counts are reported for any model. The results summarized in Table~\ref{tab:burgers0.002} show that the boosted PINN achieves the lowest MSE at $5.36 \times 10^{-2}$ and the lowest relative $L^2$ error, though the former is still above the convergence threshold. The standard PINN with Adam achieves the lowest residual norm. 

This mismatch between residual norm and MSE reflects the behavior observed in the Van der Pol example. The residual norm reflects how well the PDE is satisfied on average across all collocation points, while the MSE reflects deviation from the true solution, which for this stiff regime is concentrated in a narrow, difficult region of the domain. The standard PINN, lacking any mechanism to focus capacity on this region, distributes its residual more uniformly, yielding a low average residual norm while still missing the sharp local structure that drives the MSE. The boosted PINN, by contrast, concentrates its corrections on the most difficult region of the domain, which lowers MSE but can inflate the residual norm, since aggressive correction in a stiff zone can cause residual spikes that dominate the overall norm.

\begin{table}[ht]
    \centering
    \begin{threeparttable} 
    \begin{tabular}{lccc}
        \toprule
        & Monolithic PINN 
        & Monolithic PINN & Boosted PINN \\
        \midrule
        Optimizer                   & Adam                  & L-BFGS                
                                    & Adam + L-BFGS     \\
        Residual Norm               & $4.08 \times 10^{-2}$ & $4.58 \times 10^{-1}$ 
                                    & $1.26 \times 10^{0}$  \\
        MSE                         & $6.44 \times 10^{-1}$ & $1.89 \times 10^{-1}$ 
                                    & $3.72 \times 10^{-2}$ \\
        Relative $L^2$ Error           & $9.27 \times 10^{-1}$ & $2.72 \times 10^{-1}$ 
                                    & $5.36 \times 10^{-2}$ \\
        Training Time (s)           & --                    & --                    
                                    & --                    \\
        Number of Iterations        & --                    & --                    
                                    & --                    \\
        Learning Rate               & $10^{-2}$             & $1.0$                 
                                    & $10^{-2} \to 10^{-3}$ \\
        Training Collocation Points & $5{,}000$             & $5{,}000$             
                                    & $5{,}000$             \\
        Network Size                & $2{,}241$             & $2{,}241$             
                                    & $2{,}241$             \\
        \bottomrule
    \end{tabular}
    \caption{Model comparison for Burgers' equation ($\nu = 0.002$).}
    \label{tab:burgers0.002}
    \begin{tablenotes}
        \scriptsize
        \item \text{--} indicates that the standard PINN did not converge; iteration count and training time are therefore not reported.
    \end{tablenotes}
    \end{threeparttable}
\end{table}

\begin{figure}[H]
    \centering
    \includegraphics[width=1.0\textwidth]{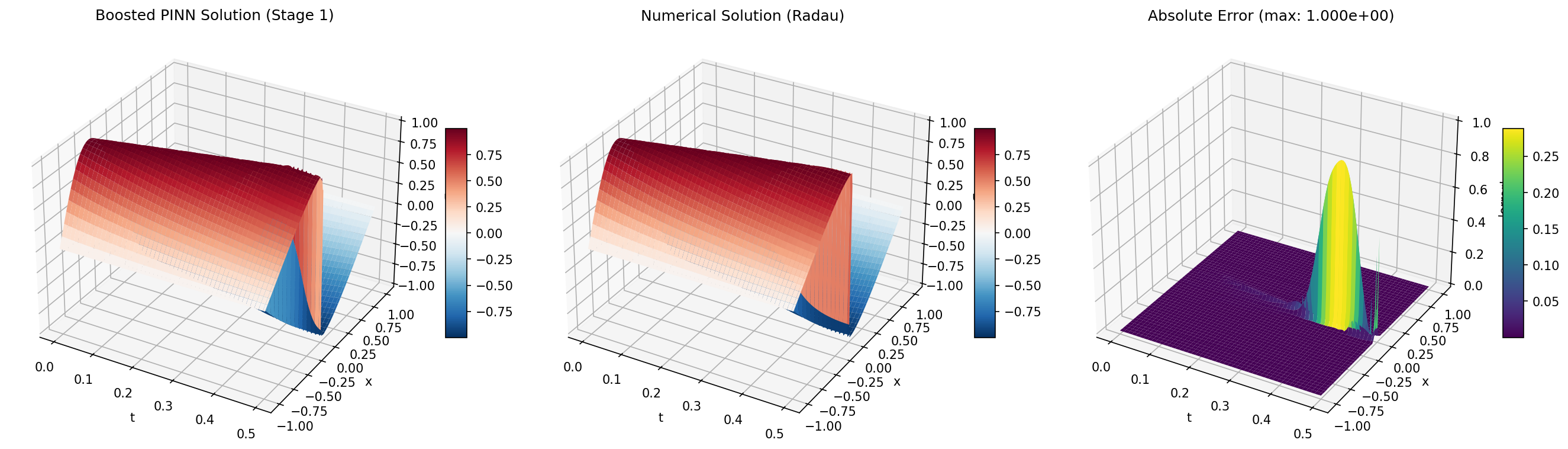}
    \caption{Boosted PINN solution vs.\ numerical solution for 
    Burgers' equation ($\nu = 0.002$).}
    \label{fig:burgers0.002_solution_fit}
\end{figure}

\begin{figure}[H]
    \centering
    \includegraphics[width=1.0\textwidth]{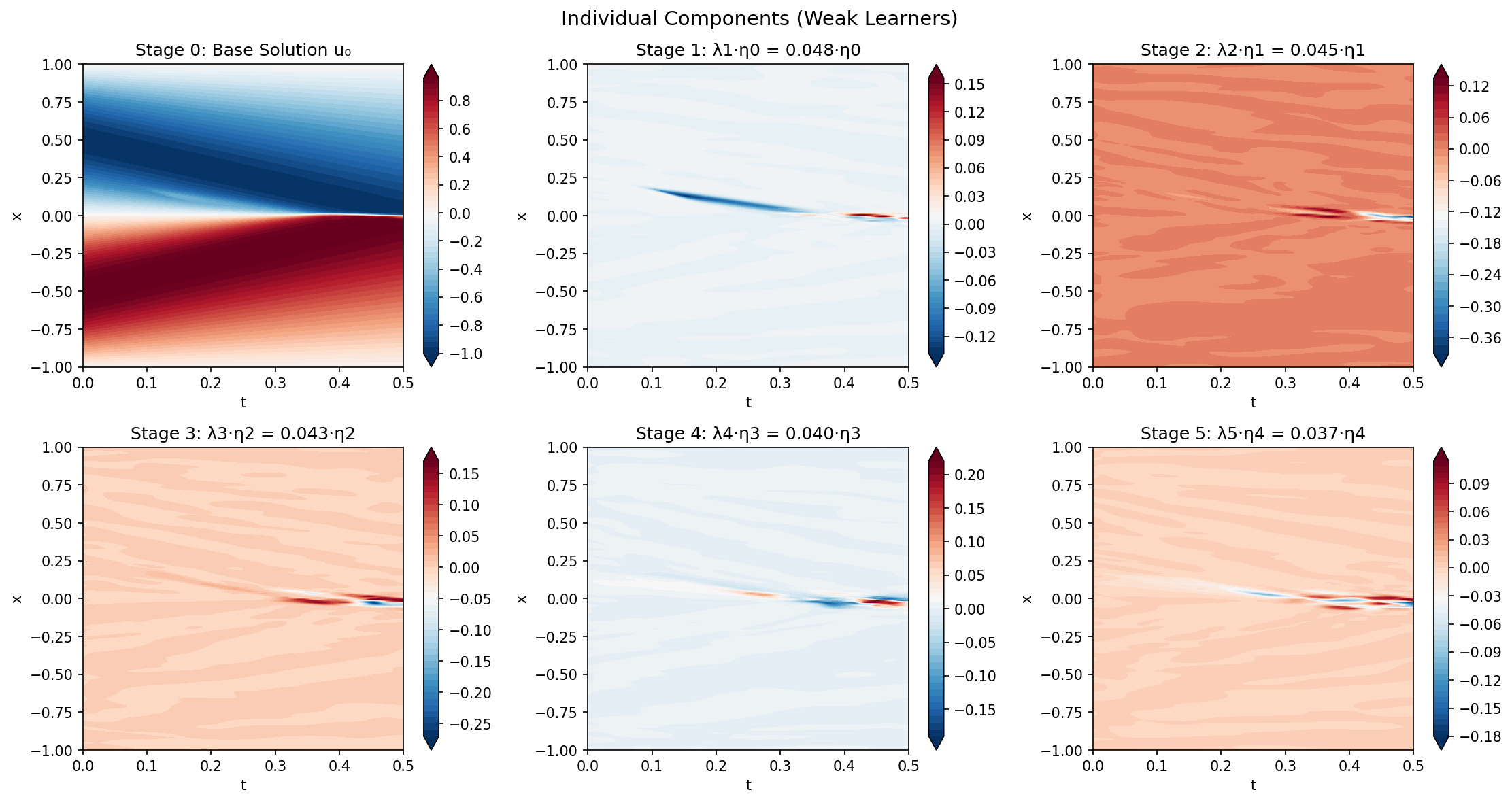}
    \caption{Boosted PINN weak learners for Burgers' equation 
    ($\nu = 0.002$).}
    \label{fig:burgers0.002_weak_learners}
\end{figure}

Stage 0 was fit using L-BFGS. For the correction stages, the first $95\%$ of epochs used Adam, while the last $5\%$ used L-BFGS. The boosted PINN solution fits well but struggles at the discontinuity, as shown in Figure~\ref{fig:burgers0.002_solution_fit} in the right-most plot. Figure~\ref{fig:burgers0.002_weak_learners} shows the weak learners across stages. Unlike in the nonstiff regime, the corrections are of much larger magnitude, indicating that the physics residual remains large throughout training. Additionally, the magnitude of the corrections does not decrease across stages, reflecting the difficulty of resolving the discontinuity, a known weakness of PINNs~\cite{Raissi2019PINNs}. Nevertheless, the boosted PINN achieves the lowest MSE and relative $L^2$ error among all three models. 

\subsubsection{Allen--Cahn with Dirichlet Boundary Conditions}
For the next example, we consider the one-dimensional Allen--Cahn equation
\begin{equation}
  \frac{\partial u}{\partial t}
  = D \frac{\partial^2 u}{\partial x^2} + u - u^3,
  \qquad (x,t) \in [-1,1] \times [0,2].
\end{equation}
The initial condition is a Gaussian profile,
$$
  u(x,0) = e^{-5x^2}, \qquad x \in [-1,1],
$$
with homogeneous Dirichlet boundary conditions,
$$
  u(-1,t) = 0, \qquad u(1,t) = 0, \qquad t \in [0,2].
$$
The diffusion parameter $D$ controls the stiffness of the equation. The smaller the value, the stiffer the equation. For this example, $D = 1.0$. Unlike the prior examples, we resample collocation points at each boosting stage for both Allen--Cahn examples. To keep parity, the standard PINN resamples the same number of times, uniformly spaced across the epoch budget.

The MSE for this specific Allen-Cahn regime was computed by comparing the PINN solution to a reference numerical solution. This reference was obtained by discretizing the spatial domain with a finite-difference Laplacian, converting the PDE into a system of ODEs, and integrating in time with a Crank–Nicolson scheme solved via Newton's method, with Dirichlet boundary conditions imposed explicitly at each step.


\paragraph{Reparameterization} We reparameterize the PINN to automatically enforce time‑dependent Dirichlet boundary conditions at $x = x_{\min}, x_{\max}$. This construction is general. For any prescribed boundary data $u_L(t)$ and $u_R(t)$, define 
\begin{equation}
    u(x_{\min}, t) = u_L(t), \quad u(x_{\max}, t) = u_R(t).
    \label{eq:allen-cahn_dir_bc_reparam}
\end{equation}
The reparameterization is given by 
\begin{equation}
u(x,t) \;=\; u_{\text{lin}}(x,t) + g(x)\,n_\theta(x,t)
\end{equation}
where
$$
u_{\text{lin}}(x,t)
= \frac{x - x_{\min}}{x_{\max} - x_{\min}}\,u_R(t)
+ \frac{x_{\max} - x}{x_{\max} - x_{\min}}\,u_L(t),
$$
$$
g(x) = (x - x_{\min})(x_{\max} - x),
$$
and $n_\theta(x,t)$ denotes the neural network output with parameters $\theta$. Note that, for our specific case, the reparameterization simplifies to
\begin{equation}
    u(x,t) \;=\; g(x)\,n_\theta(x,t),
\end{equation}
since $u_L(t) = u_R(t) = 0$.

This guarantees the Dirichlet conditions. At $x = x_{\min}$, we have $u_{\text{lin}}(x_{\min}, t) = u_L(t)$ and $g(x_{\min}) = 0$, so $u(x_{\min}, t) = u_L(t)$. Similarly, at $x = x_{\max}$, $u_{\text{lin}}(x_{\max}, t) = u_R(t)$ and $g(x_{\max}) = 0$, so $u(x_{\max}, t) = u_R(t)$, matching \eqref{eq:allen-cahn_dir_bc_reparam}.

We also reparameterize the weak learners' networks so that their corrections vanish at the boundary, preserving the boundary conditions already satisfied by stage 0. We reparameterize as follows: Define
\begin{equation}
    c(x,t) = g(x)\,n_\theta(x,t),
\end{equation}
where $g(x)$ is as defined above.
At the boundaries:
$$g(x_{\min}) = (x_{\min} - x_{\min})(x_{\max} - x_{\min}) = 0,$$
$$g(x_{\max}) = (x_{\max} - x_{\min})(x_{\max} - x_{\max}) = 0,$$
hence
$$c(x_{\min},t) = 0, \quad c(x_{\max},t) = 0 \quad \forall t.$$
In the interior $x \in (x_{\min}, x_{\max})$, $g(x) \neq 0$, so the network can express arbitrary corrections (scaled by $g$) while always satisfying
$$c(x_{\min},t) = c(x_{\max},t) = 0.$$

\begin{figure}[H]
    \centering
    \includegraphics[width=1.0\textwidth]{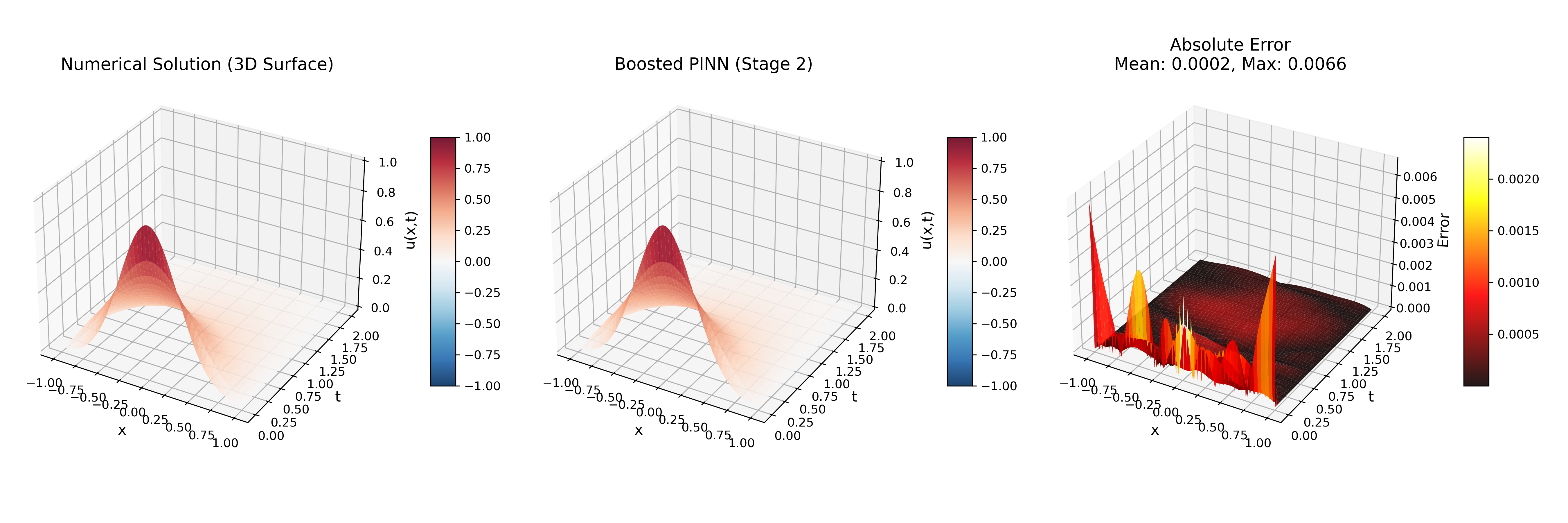}
    \caption{Boosted PINN solution vs.\ numerical solution for the 
    Allen--Cahn equation ($D = 1.0$).}
    \label{fig:allencahn1.0_solution_fit}
\end{figure}

\paragraph{Results, \boldmath{$D = 1.0$}}
As shown in Table~\ref{tab:allencahn1.0} and 
Figure~\ref{fig:allencahn1.0_solution_fit}, the residual norm, MSE, and relative $L^2$ error of the standard PINN using L-BFGS are the lowest, at $5.46 \times 10^{-3}$, $8.10 \times 10^{-5}$, and $4.21 \times 10^{-4}$, respectively. For these same performance metrics, the boosted PINN's are one order of magnitude worse, at $1.88 \times 10^{-2}$, $3.68 \times 10^{-4}$, and $1.91 \times 10^{-3}$. The boosted PINN, however, converges the quickest, taking 2.78 seconds and 500 epochs, on average. The standard PINN also takes 500 epochs to converge but converges more slowly, requiring 3.51 seconds.

\begin{table}[H]
    \centering
    \begin{threeparttable}
    \begin{tabular}{lccc}
        \toprule
        & Monolithic PINN 
        & Monolithic PINN & Boosted PINN \\
        \midrule
        Optimizer             & Adam                  & L-BFGS                
                              & Adam                  \\
        Residual Norm         & $1.75 \times 10^{-2}$ & $5.46 \times 10^{-3}$ 
                              & $1.88 \times 10^{-2}$ \\
        MSE                   & $2.50 \times 10^{-3}$ & $8.10 \times 10^{-5}$ 
                              & $3.68 \times 10^{-4}$ \\
        Relative $L^2$ Error     & $1.30 \times 10^{-2}$ & $4.21 \times 10^{-4}$ 
                              & $1.91 \times 10^{-3}$ \\
        Training Time (s)     & $3.51$                & $24.73$              
                              & $2.78$                \\
        Number of Iterations  & $500$                 & $4{,}062$\tnote{*}             
                              & $500$             \\
        Learning Rate         & $10^{-2}$             & $1.0$                 
                              & $10^{-2}$             \\
        Collocation Points    & $5{,}000$             & $5{,}000$             
                              & $5{,}000$             \\
        Network Size          & $4{,}417$             & $4{,}417$            
                              & $4{,}417$             \\
        \bottomrule
    \end{tabular}
    \caption{Model comparison for the Allen--Cahn equation ($D = 1.0$).}
    \label{tab:allencahn1.0}
    \begin{tablenotes}
        \scriptsize
        \item[*] L-BFGS ran for $4{,}062$ function evaluations across $500$ epochs.
    \end{tablenotes}
    \end{threeparttable}
\end{table}

\begin{figure}[H]
    \centering
    \includegraphics[width=1.0\textwidth]{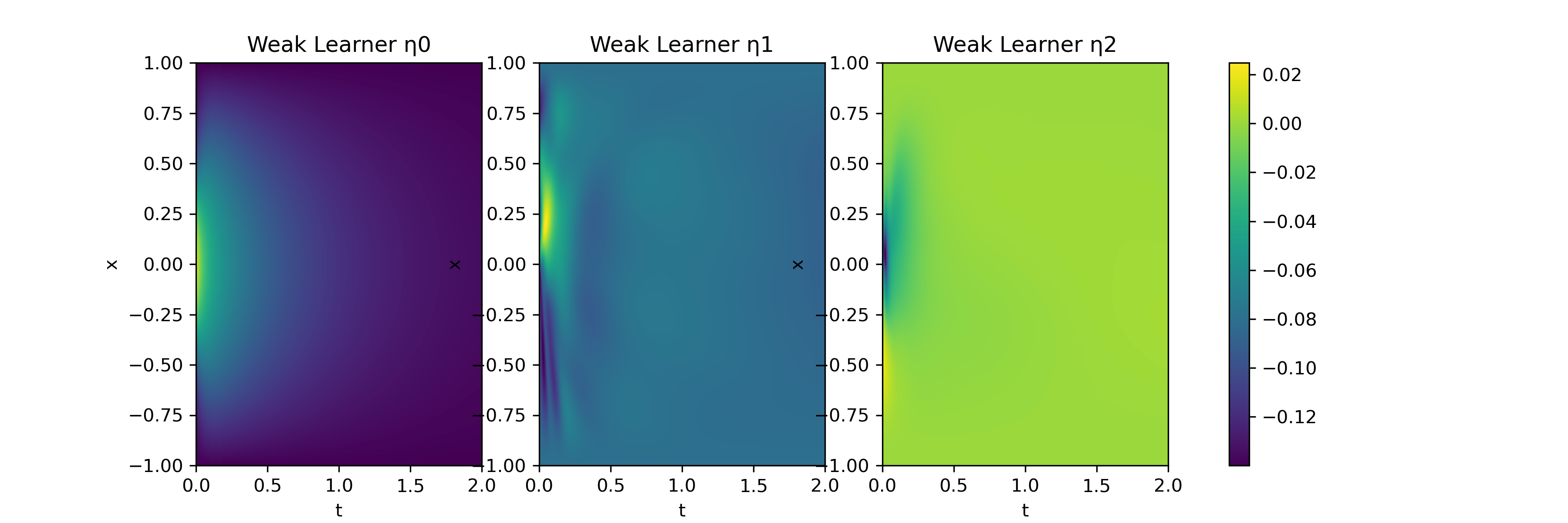}
    \caption{Boosted PINN weak learners for the Allen--Cahn equation 
    ($D = 1.0$).}
    \label{fig:allencahn1.0_weak_learners}
\end{figure}

The Allen--Cahn equation with Dirichlet boundary conditions and a large $D$ value is smooth and poses little difficulty for PINNs. As such, we only use three stages for the boosted PINN; additional stages would only add noise, corrupting the solution. The right-hand plot of Figure~\ref{fig:allencahn1.0_solution_fit} shows the absolute error. At the boundaries $x_{\min}$ and $x_{\max}$, the error is exactly zero, as a result of the reparameterization of stage 0. The error remains zero even after subsequent corrections because the weak learner corrections are also reparameterized to vanish at the boundaries, preserving the boundary conditions already satisfied by stage 0. Figure~\ref{fig:allencahn1.0_weak_learners} shows the corrections concentrated at the initial condition, since most of the error occurs at this time point.

\subsubsection{Allen--Cahn Equation with Periodic Boundary Conditions}
For this example, we switch from Dirichlet to periodic boundary conditions. Periodic boundary conditions combined with a small diffusion coefficient $D = 10^{-4}$ cause the solution to develop sharp interfaces between phases, inducing stiffness.
We consider the Allen--Cahn equation
\begin{equation}
    \frac{\partial u}{\partial t}
    = D\,\frac{\partial^2 u}{\partial x^2} + u - u^3,
    \qquad
    x \in [-1,\,1],\quad t \in [0,\,3],
    \label{eq:allen_cahn_periodic}
\end{equation}
where $D$ is the diffusion coefficient governing interface width. The stable equilibria of the reaction term $u - u^3$ are $u = \pm 1$, and for small $D$, the solution develops sharp interfaces between these two phases.
The initial condition is
\begin{equation}
    u(x,0) = x^2 \cos(\pi x).
    \label{eq:allen_cahn_periodic_ic}
\end{equation}
Periodic boundary conditions are imposed on the domain boundaries 
$x = -1$ and $x = 1$:
\begin{equation}
    u(-1,\,t) = u(1,\,t),
    \qquad
    \frac{\partial u}{\partial x}\bigg|_{x=-1}
    =
    \frac{\partial u}{\partial x}\bigg|_{x=1},
    \qquad
    t \in [0,\,3].
    \label{eq:allen_cahn_periodic_bc}
\end{equation}

The MSE of the boosted PINN was computed by comparing the PINN solution to a Fourier spectral reference solution. The reference is solved on a 512-point Fourier grid in space (method of lines) with adaptive implicit (Radau) time-stepping, then interpolated to a 256-point uniform spatial grid × 201-point time grid for the actual PINN comparison.

Also note that this example differs in the sampling strategy used in prior examples (Section~\ref{section:sampling_strategy}). Here, we sample a new set of collocation points for each stage. To make comparisons equivalent between the boosted PINN and standard PINN, we have the standard PINN resample an equal number of times, spaced uniformly across the epoch budget. For example, if we have a budget of 10,000 epochs and ten stages, then every 1,000 epochs, the standard PINN gets a new set of collocation points.

\paragraph{Reparameterization}
We reparameterize both stage 0 and the correction stages to remove the boundary condition loss from the PINN's total loss. Let $L = x_{\max} - x_{\min} = 2$ be the period of the domain $[x_{\min}, x_{\max}] = [-1, 1]$. For a spatial coordinate $x$, define the phase angle
\begin{equation}
    \varphi(x) = \frac{2\pi\,(x - x_{\min})}{L}
              = \pi(x + 1),
    \qquad \varphi \in [0,\, 2\pi].
    \label{eq:phase_angle}
\end{equation}

The two-dimensional Fourier embedding is then
\begin{equation}
    \boldsymbol{\phi}(x)
    =
    \begin{pmatrix}
        \cos\varphi(x) \\[4pt]
        \sin\varphi(x)
    \end{pmatrix},
    \label{eq:fourier_embed}
\end{equation}
and the augmented network input is
\begin{equation}
    \mathbf{z}(x,t)
    =
    \begin{pmatrix}
        \cos\bigl(\pi(x+1)\bigr) \\[4pt]
        \sin\bigl(\pi(x+1)\bigr) \\[4pt]
        t
    \end{pmatrix}
    \in \mathbb{R}^3.
    \label{eq:network_input}
\end{equation}

The stage 0 approximation is defined as
\begin{equation}
    u^{(0)}(x, t)
    = \phi_0\!\bigl(\mathbf{z}(x,t);\,\theta_0\bigr),
    \label{eq:stage0_reparam}
\end{equation}
where $\phi_0:\mathbb{R}^3\to\mathbb{R}$ is a fully connected neural network and $\theta_0$ are its trainable parameters.

At $x = x_{\min} = -1$ and $x = x_{\max} = 1$, the phase angle~\eqref{eq:phase_angle} takes the values $\varphi(-1) = 0$ and $\varphi(1) = 2\pi$. Since $\cos$ and $\sin$ are $2\pi$-periodic,
\begin{equation}
    \mathbf{z}(-1, t)
    =
    \begin{pmatrix} 1 \\ 0 \\ t \end{pmatrix}
    =
    \mathbf{z}(1, t),
    \label{eq:embed_equality}
\end{equation}
so that $u^{(0)}(-1,t) = u^{(0)}(1,t)$ for all $t$.

By the chain rule,
\begin{equation}
    \frac{\partial u^{(0)}}{\partial x}
    =
    \frac{\partial \phi_0}{\partial \mathbf{z}}
    \cdot
    \frac{d\mathbf{z}}{d x}
    =
    \pi\,\frac{\partial \phi_0}{\partial \mathbf{z}}
    \cdot
    \begin{pmatrix}
        -\sin\varphi \\[4pt]
         \cos\varphi \\[4pt]
         0
    \end{pmatrix}.
    \label{eq:deriv_chain}
\end{equation}

Because $(-\sin 0,\,\cos 0) = (0,\,1) = (-\sin 2\pi,\,\cos 2\pi)$, the input Jacobian $d\mathbf{z}/dx$ is identical at $x = -1$ and $x = 1$. With~\eqref{eq:embed_equality}, this guarantees
\begin{equation}
    \left.\frac{\partial u^{(0)}}{\partial x}\right|_{x=-1}
    =
    \left.\frac{\partial u^{(0)}}{\partial x}\right|_{x=1},
    \label{eq:deriv_periodic}
\end{equation}
satisfying the full periodic boundary condition~\eqref{eq:allen_cahn_periodic_bc} exactly.

Each correction $h_k$ at boosting stage $k$ uses the identical embedding:
\begin{equation}
    h_k(x,t)
    = \phi_k\!\bigl(\mathbf{z}(x,t);\,\theta_k\bigr),
    \label{eq:correction_reparam}
\end{equation}
so that the periodic conditions $h_k(-1,t) = h_k(1,t)$ and $\partial_x h_k|_{x=-1} = \partial_x h_k|_{x=1}$ hold for any choice of $\theta_k$. Since every correction satisfies the periodic boundary conditions by construction, so does their sum, and the composite loss no longer requires a boundary term. Only the PDE residual and initial-condition terms remain.

\paragraph{Results, \boldmath$D = 10^{-4}$}
Choosing $D = 10^{-4}$ imbues stiffness into the equation. The standard PINNs with both Adam and L-BFGS do not converge, as their MSEs do not fall below $< 10^{-2}$ within the given epoch budget. Their residual norm and MSE reported in Table~\ref{tab:allen_cahn_periodic} are the best values achieved before termination. The boosted PINN consistently converges across independent runs, using $16{,}933$ epochs and $499.44$ seconds, on average. The boosted PINN achieved an MSE of $ 7.10 \times 10^{-3}$, a residual norm of $2.37 \times 10^{-3}$ and a relative $L^2$ error of $1.22 \times 10^{-2}$, all of which are lower than the standard PINNs. These are the best values achieved before termination. This result is consistent with prior observations: for stiff or complex equations, the standard PINN fails to converge while the boosted PINN is successful.

\begin{table}[H]
    \centering
    \begin{threeparttable}
    \begin{tabular}{lccc}
        \toprule
         & \textbf{Monolithic PINN} & \textbf{Monolithic PINN} 
         & \textbf{Boosted PINN} \\
        \midrule
        Optimizer            & Adam                  & L-BFGS                
                             & Adam                  \\
        Residual Norm        & $2.40 \times 10^{-2}$ & $3.66 \times 10^{-3}$ 
                             & $2.37 \times 10^{-3}$ \\
        MSE                  & $1.89 \times 10^{-1}$ & $1.41 \times 10^{-2}$ 
                             & $7.10 \times 10^{-3}$ \\
        Relative $L^2$ Error    & $3.24 \times 10^{-1}$ & $2.60 \times 10^{-2}$ 
                             & $1.22 \times 10^{-2}$ \\
        Training Time (s)    & --                    & --                    
                             & $499.44$              \\
        Number of Iterations & --                    & --       
                             & $16{,}933$            \\
        Learning Rate        & $10^{-2} \to 1.91 \times 10^{-8}$ 
                             & $1.0$                 
                             & $10^{-2} \to 5.0 \times 10^{-3}$ \\
        Collocation Points   & $5{,}000$             & $5{,}000$             
                             & $5{,}000$             \\
        Network Size         & $8{,}641$             & $8{,}641$            
                             & $8{,}641$             \\
        \bottomrule
    \end{tabular}
    \caption{Model comparison for the Allen--Cahn equation with periodic 
    boundary conditions ($D = 10^{-4}$).}
    \label{tab:allen_cahn_periodic}
    \begin{tablenotes}
        \scriptsize
        \item[--] Monolithic PINN (Adam and L-BFGS) did not 
        converge to MSE $< 10^{-2}$; training time and iteration count 
        are therefore not reported. The residual norm, MSE, and relative $L^2$ error reported are the best values achieved before termination.
    \end{tablenotes}
    \end{threeparttable}
\end{table}

\begin{figure}[ht]
    \centering
    \includegraphics[width=1.0\textwidth]{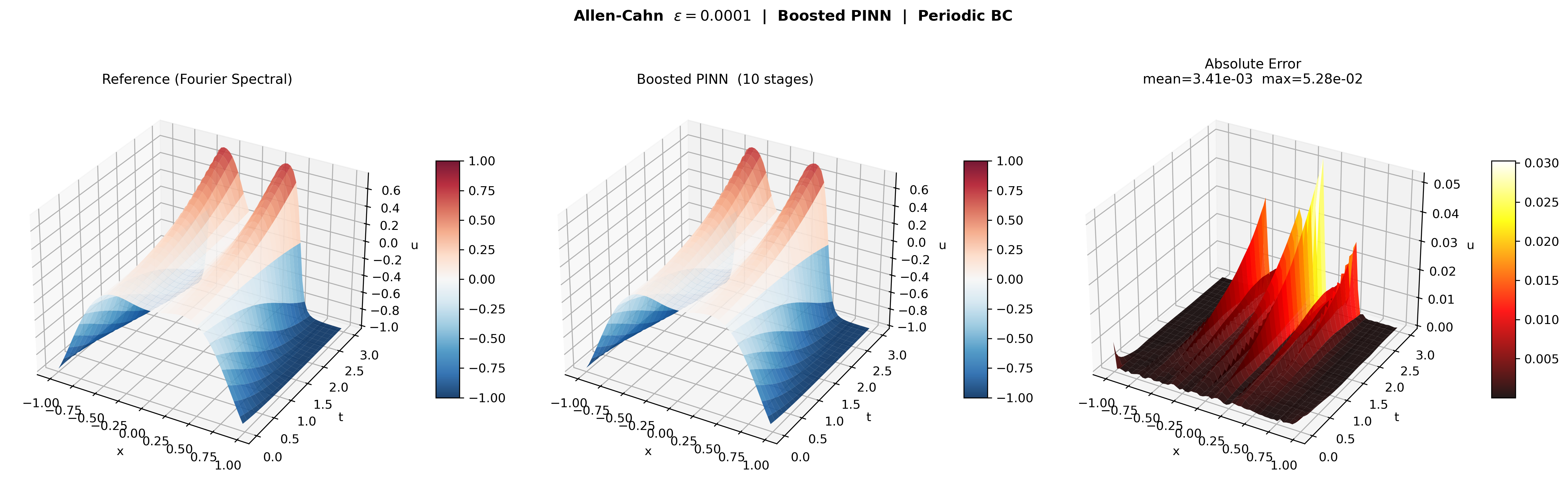}
    \caption{Boosted PINN solution vs.\ numerical solution for the 
    Allen--Cahn equation ($D = 10^{-4}$).}
    \label{fig:allencahn0.0001_solution_fit}
\end{figure}

\begin{figure}[ht]
    \centering
    \includegraphics[width=1.0\textwidth]{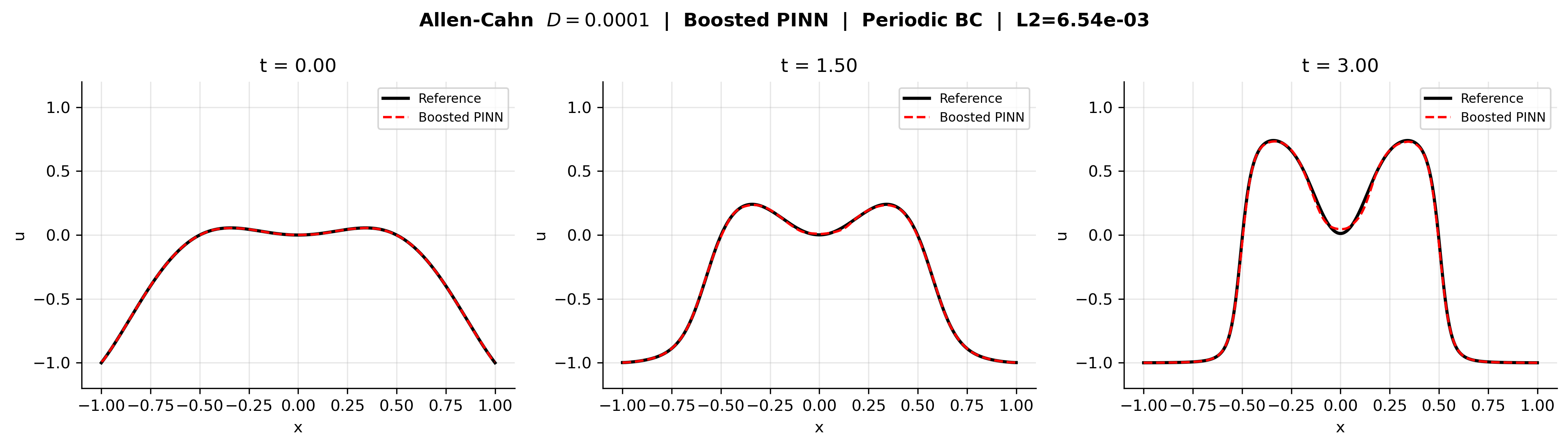}
    \caption{Boosted PINN solution slices at selected time points for 
    the Allen--Cahn equation ($D = 10^{-4}$).}
    \label{fig:allencahn0.0001_solution_slices}
\end{figure}

Figures~\ref{fig:allencahn0.0001_solution_fit} and~\ref{fig:allencahn0.0001_solution_slices} show the boosted PINN solution as a 3D plot and solution slices at selected time points. The solution slices show that the boosted PINN performs well at early time points but struggles somewhat at later time points, consistent with the increasing stiffness of the problem over time. Note that the absolute error at the boundaries $x_{\min}$ and $x_{\max}$ is exactly zero due to the reparameterization. In total, ten correction stages were used.

\begin{figure}[H]
    \centering
    \includegraphics[width=1.0\textwidth]{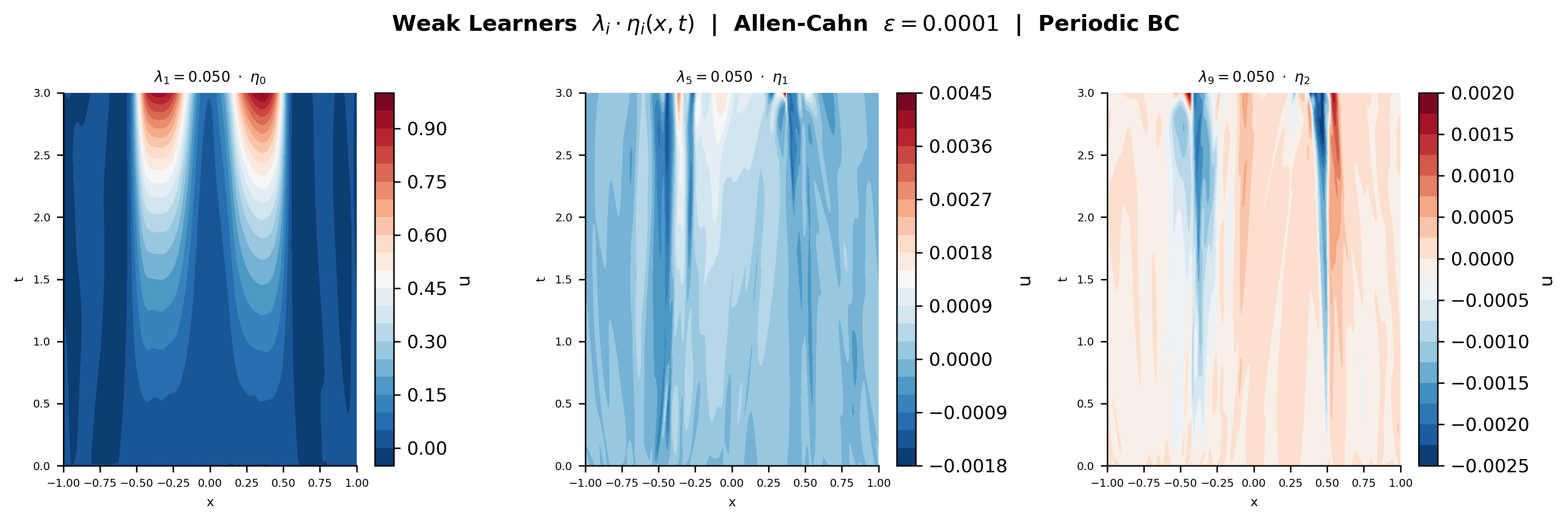}
    \caption{Boosted PINN weak learners for the Allen--Cahn equation 
    ($D = 10^{-4}$).}
    \label{fig:allencahn0.0001_weak_learners}
\end{figure}

Figure~\ref{fig:allencahn0.0001_weak_learners} shows the weak learners across stages. As the number of stages increases, the PDE residual decreases and the magnitude of each correction decreases accordingly, as indicated by the decreasing scale shown on the vertical axis across stages. This is the ideal boosting behavior described in Section~\ref{subsection:projected_descent_and_optimality}.

\section{Ablation Study}
\label{sec:ablation_study}
\paragraph{Transfer Learning Ablation}
We present results for the transfer learning ablation experiment, which examines the effect of removing transfer learning between boosting stages. The hypothesis is that removing transfer learning will cause the MSE to increase. We selected three examples spanning different problem types: an ODE (Duffing), a coupled ODE (Lotka--Volterra), and a PDE (Burgers'). We generated a paired dataset using ten independent seeds; for a given seed, we ran the boosted PINN with and without transfer learning.

From Table~\ref{tab:ablation_transfer}, the effect of transfer learning is statistically significant by the Wilcoxon signed-rank test for the Lotka--Volterra and Burgers' examples, but not for the Duffing example ($p = 0.3125$). For the Duffing example, the difference is not statistically significant, likely due to high variance across seeds, though a practical difference in MSE is observed ($2.04 \times 10^{-3}$ vs.\ $8.90 \times 10^{-3}$). For the Lotka--Volterra system, transfer learning is critical: without it, the MSE is orders of magnitude worse ($p = 0.0042$). For Burgers', transfer learning gives a modest but significant improvement ($p = 0.0420$).

\begin{table}[H]
\centering
    \begin{threeparttable}
        \begin{tabular}{lccc}
            \toprule
            Problem & MSE (Transfer) & MSE (No Transfer) & Wilcoxon $p$ \\
            \midrule
            Duffing         & $2.04 \times 10^{-3}$ & $8.90 \times 10^{-3}$ 
                            & $0.3125$              \\
            Lotka--Volterra & $1.76 \times 10^{-3}$ & $7.91 \times 10^{-1}$ 
                            & $0.0042^{**}$         \\
            Burgers'        & $1.77 \times 10^{-4}$ & $2.91 \times 10^{-4}$ 
                            & $0.0420^{*}$          \\
            \bottomrule
        \end{tabular}
        \caption{Effect of transfer learning on MSE.}
        \label{tab:ablation_transfer}
        \begin{tablenotes}
            \scriptsize
            \item[$^{*}$] $p < 0.05$. 
            \item[$^{**}$] $p < 0.01$.
        \end{tablenotes}
    \end{threeparttable}
\end{table}

\section{Sensitivity Analysis}
\label{sec:sensitivity_analysis
}

In this section, we report the results of a sensitivity analysis of the boosted PINN, varying the number of neurons per layer and the number of stages. Recall that the hidden dimension of the weak learners is uniform. We report findings for two examples: a coupled ODE system (Lotka-Volterra) and a stiff PDE (Allen-Cahn with periodic boundary conditions). We report three metrics. The first is the MSE achieved at the final stage of the boosted PINN. The second is the time until convergence, and the last is the number of epochs until convergence. Each metric is averaged over five seeds.

Our hypotheses are as follows:
\begin{enumerate}
    \item As the number of stages increases, the MSE should decrease, eventually plateauing~\citep{chaudhry-2026}.
    \item As the number of neurons increases, the MSE should decrease but only up to a point because if the size of the weak learners is too large, they begin to resemble monolithic PINNS, which we have shown (Section \ref{sec:experiments}) perform worse than fitting small learners sequentially.
\end{enumerate}

\begin{figure}[H]
    \centering
    \includegraphics[width=0.85\textwidth]{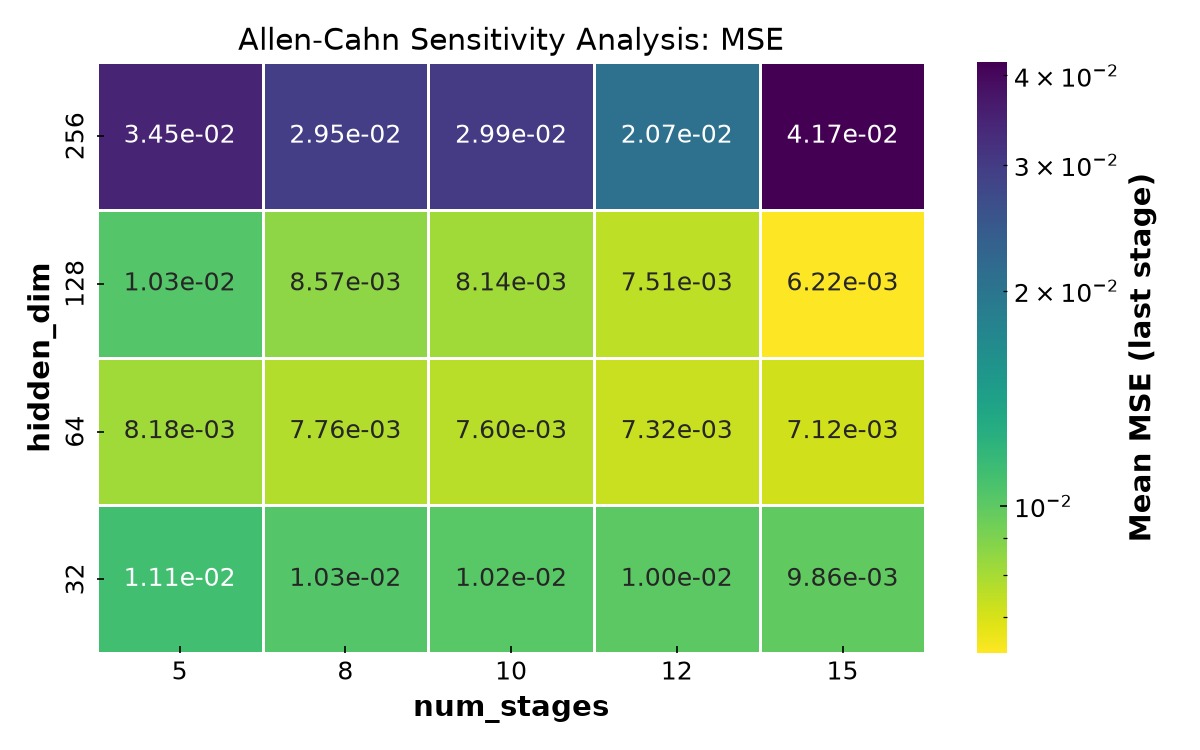}
    \caption{MSE Sensitivity - Allen-Cahn}
    \label{fig:allen-cahn_heatmap_l2_error}
\end{figure}

\paragraph{Lotka-Volterra} The results for Lotka-Volterra are consistent with our hypotheses: increasing model size and the number of stages both reduce the MSE. Notably, when a weak learner lacks sufficient capacity to approximate the functional gradient, increasing the number of stages fails to improve performance, as shown in the last row of \ref{fig:allen-cahn_heatmap_l2_error}. In this regime, the model cannot adequately descend the space of functions, and convergence stalls.

\begin{figure}[H]
    \centering
    \includegraphics[width=0.85\textwidth]{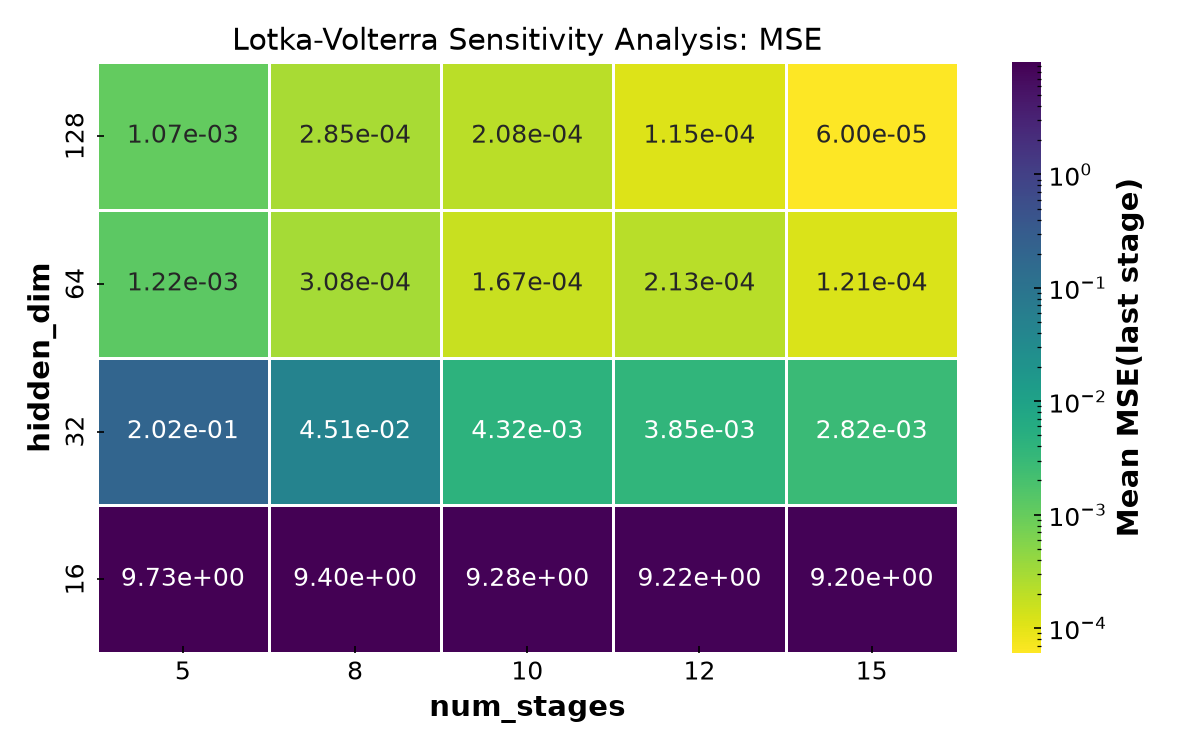}
    \caption{MSE Sensitivity - Lotka-Volterra}
    \label{fig:lk_heatmap_l2_error}
\end{figure}

Next, we examine the sensitivity of convergence metrics (Figures \ref{fig:lk_heatmap_l2_error} and \ref{fig:lk_heatmap_epoch_to_cross}). The left-hand plot reports the metric averaged over seeds that converged, conditional on convergence. The right-hand grayscale plot reports the proportion of seeds that converged for each configuration, serving as a measure of training stability.

Holding the number of stages fixed, the number of epochs required for convergence decreases monotonically as the size of the weak learner increases. Training stability follows the same trend. This is consistent with our hypotheses: a larger weak learner is more expressive and can more accurately estimate the functional gradient. For a fixed hidden dimension, we observe no meaningful variation in the number of epochs to convergence across the number of stages. Each stage is trained independently, and its epoch budget is therefore unaffected by the number of stages preceding it. In terms of wall-clock time to convergence, we similarly observe little variation across stages for a fixed hidden dimension, but substantial variation across the hidden dimension axis. As the number of neurons increases, per-epoch computational cost grows, increasing total training time.

\begin{figure}[H]
    \centering
    \includegraphics[width=1.0\textwidth]{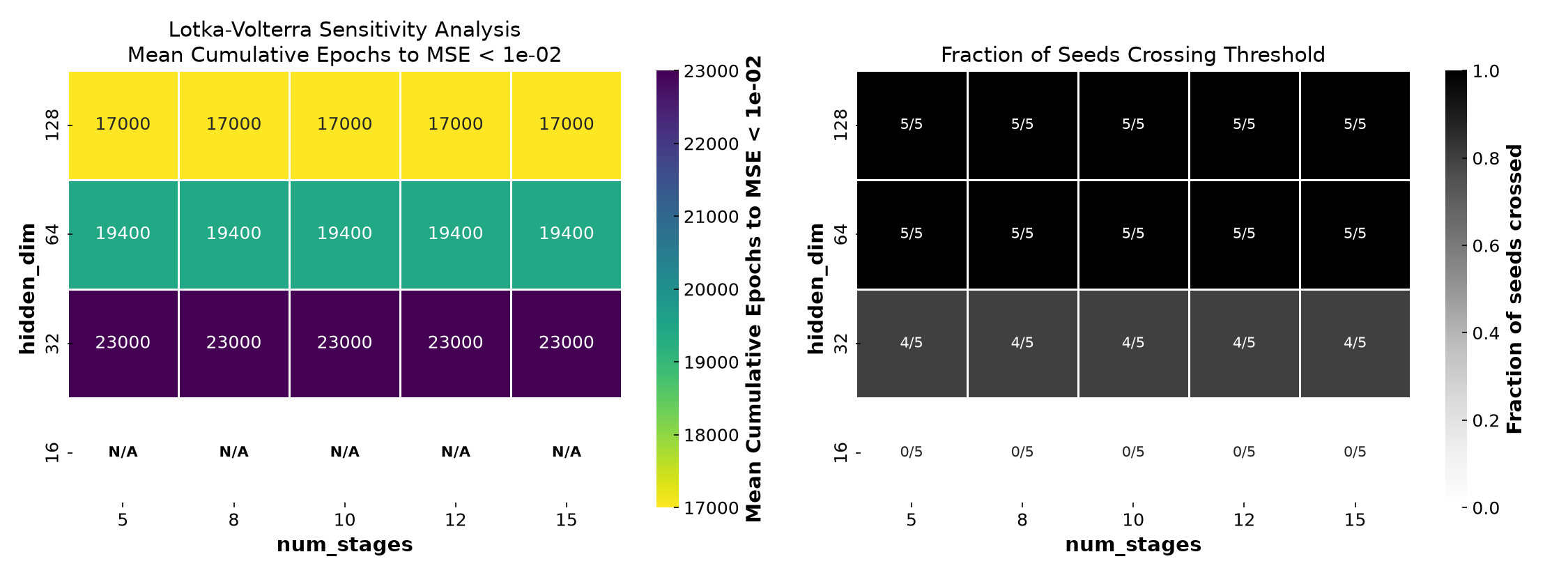}
    \caption{Crossing Epoch Sensitivity - Lotka-Volterra}
    \label{fig:lk_heatmap_epoch_to_cross}
\end{figure}

\begin{figure}[H]
    \centering
    \includegraphics[width=1.0\textwidth]{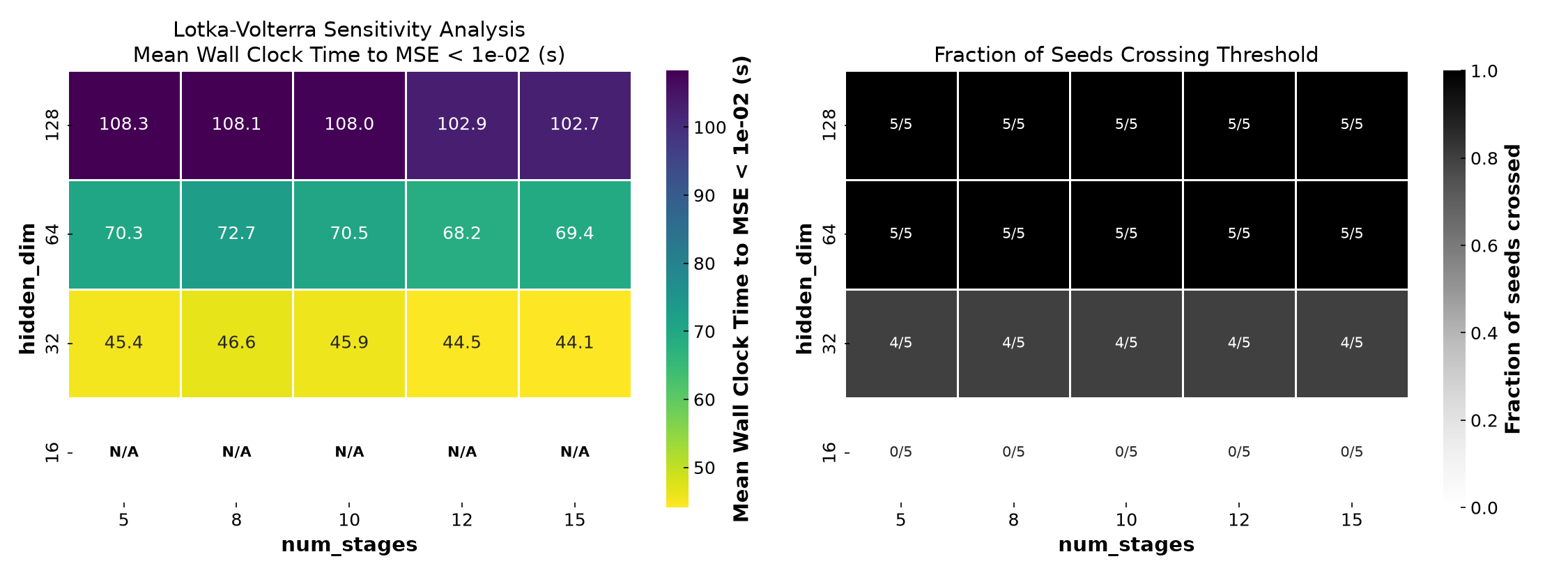}
    \caption{Wall Clock Sensitivity - Lotka-Volterra}
    \label{fig:lk_heatmap_epoch_to_cross}
\end{figure}

\paragraph{Allen-Cahn} We perform the same sensitivity analysis on the Allen-Cahn PDE under a stiff regime. As shown in Figure \ref{fig:allen-cahn_heatmap_l2_error}, most of the variation in MSE arises from the hidden dimension rather than the number of stages. However, unlike the ODE example, increasing the number of neurons improves performance only up to a point, beyond which performance degrades. For this example, a hidden dimension between 64 and 128 performs best; at 256, performance degrades to the same order of magnitude as at 32. The observed degradation in the large weak learner size may be attributable to the same spectral bias \citep{pmlr-v97-rahaman19a}, observed in standard PINNs, which would render each stage's estimate of the functional gradient increasingly inaccurate. 

The crossing metrics are non-monotonic for the stiff PDE case. For the stiff PDE case, the crossing metrics behave non-monotonically with respect to model size. Stable training is confined to a specific region of the parameter space spanned by the number of stages and hidden dimension: specifically, a hidden dimension of 64, or a hidden dimension of 128 combined with a large number of stages. Within this stable region, epochs-to-convergence are lowest at a hidden dimension of 64, while wall-clock time to convergence increases with model size.

Taken together, the sensitivity analysis shows that for non-stiff systems, increasing model size and the number of stages monotonically improves performance. For stiff systems, by contrast, selecting an appropriately sized weak learner is critical: if the weak learner is too small, it fails to accurately estimate the functional gradient; if it is too large, it behaves like a monolithic PINN and suffers from the same ill-conditioning, spectral bias, and unstable training dynamics mentioned in the introduction (Section \ref{section:introduction}).

\begin{figure}[H]
    \centering
    \includegraphics[width=1.0\textwidth]{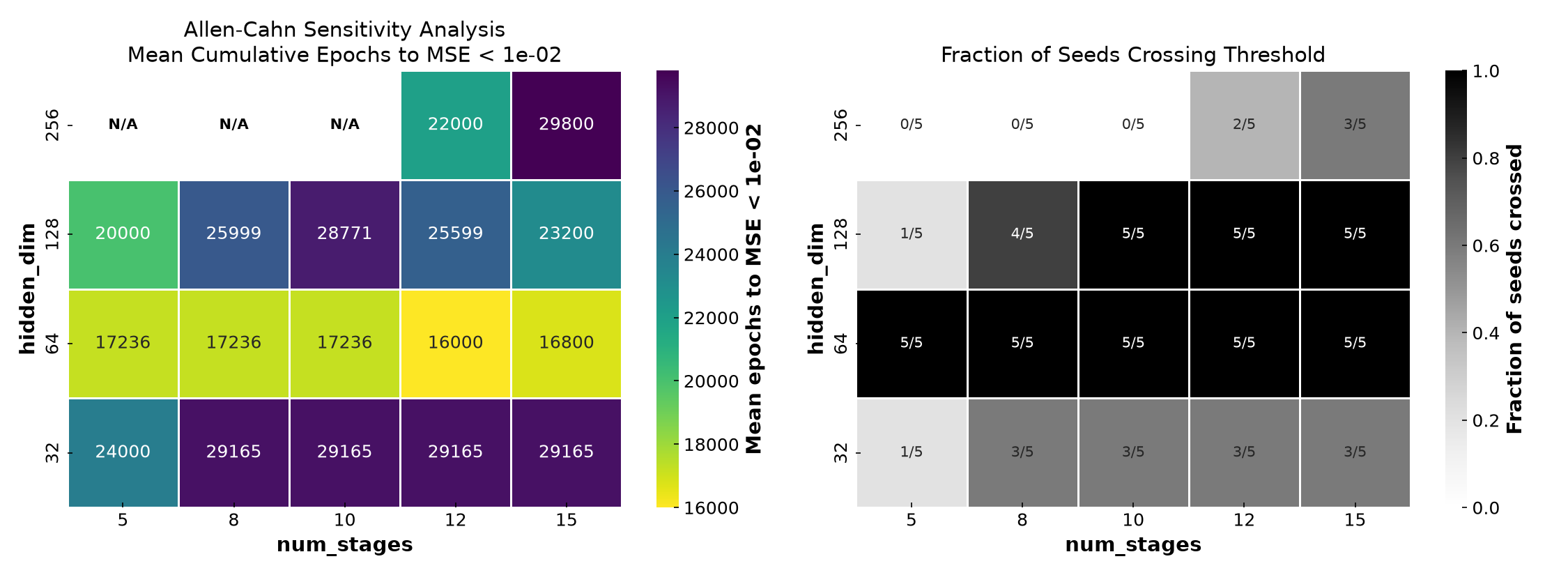}
    \caption{}
    \label{fig:allen-cahn_heatmap_epoch_to_cross}
\end{figure}
\begin{figure}[H]
    \centering
    \includegraphics[width=1.0\textwidth]{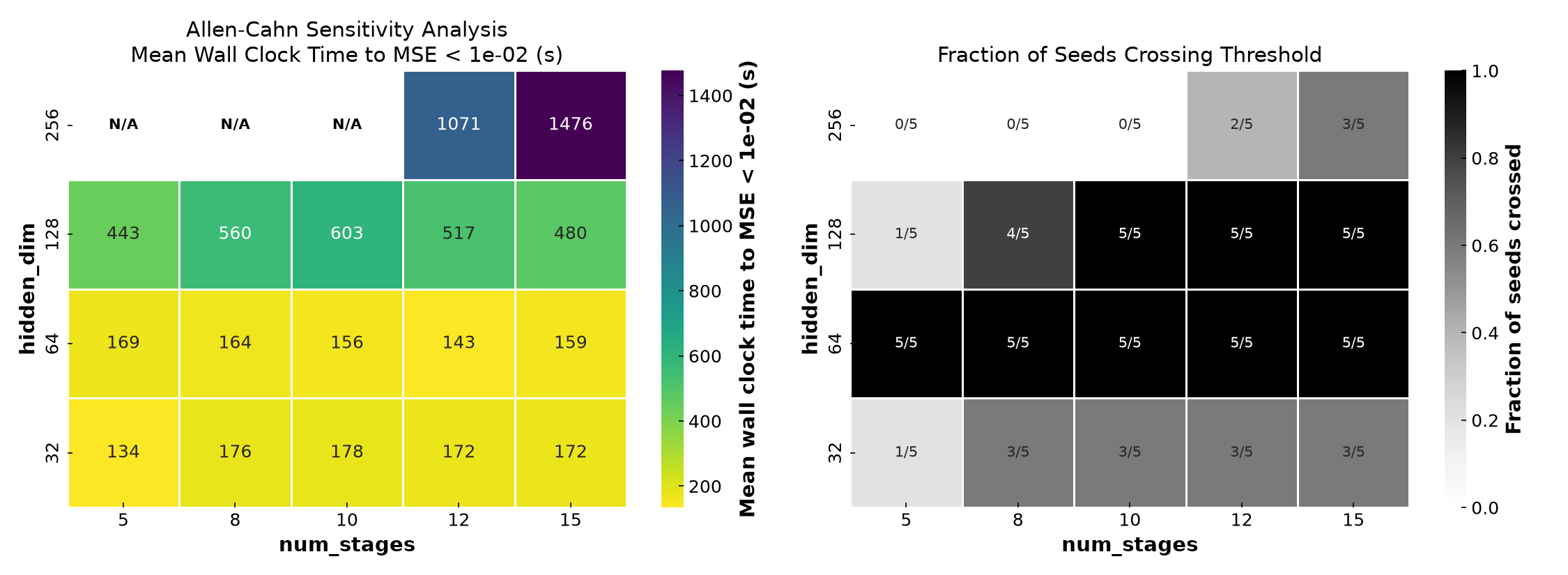}
    \caption{}
    \label{fig:allen-cahn_heatmap_epoch_to_cross}
\end{figure}

\section{Discussion}
\label{sec:discussion}
The proposed framework separates nonlinear refinement into a sequence of low-dimensional variational problems. Unlike classical boosting, each stage minimizes the full nonlinear residual, preserving operator structure. The approach enables second-order optimization while maintaining a principled geometric interpretation. We also showed that transfer learning across the boosting stages improves model performance.

Our experiments show that for smooth, nonstiff problems, the standard PINN with L-BFGS outperforms the boosted PINN, as in the case of the viscous Burgers' equation and Allen--Cahn in the nonstiff regime. Even in cases where the boosted PINN falls short in terms of accuracy, it outperforms in terms of convergence, measured in time and epochs. Furthermore, for stiff or vector-valued problems, the boosted PINN converges with lower error and, in some cases, converges to a solution when the standard PINN could not converge, as with the NRD and Van der Pol ODEs and the Allen--Cahn PDE. 

\paragraph{Limitations}
There is a level of stiffness that the boosted PINN cannot handle, as shown for Van der Pol with $\mu = 4.0$ and Burgers' with $\nu = 0.002$. Although not shown, the boosted PINN also suffers from long temporal problems, similar to standard PINNs. If one extends beyond one cycle in the Lotka--Volterra example, all models fail to converge. Furthermore, our examples only use second-order optimizers for nonstiff ODEs. Implementing second-order optimizers for nonstiff PDEs and stiff equations remains future work.

Despite these limitations, the results make clear that boosting as an optimization framework is a valid approach to training PINNs. We successfully showed that boosting works for ODE IVPs and BVPs, both stiff and nonstiff, for coupled ODEs, and for PDEs in both stiff and nonstiff regimes. Our results reflect the theory: the weak learners learned non-trivial approximations to the projected functional gradient, and the ensemble converged toward the true solution with each additional weak learner. Boosting provides an alternative that uses far fewer parameters and less compute, while offering faster convergence times.

\section{Conclusion}
We introduced a variational boosting formulation of PINNs, grounded in existence and uniqueness theory for the underlying residual-minimization problem. By performing descent in function space through restricted variational minimization, the method provides stability, modular refinement, and practical second-order solvability. This perspective bridges variational PDE theory, functional gradient methods, and neural operator learning.

Since the weak learners themselves are small, we were able to successfully implement and train two custom second-order optimizers, namely conjugate gradient and Newton. Both optimizers improve the training time and the error for the boosted PINN. Across a range of ODE, coupled ODE, and PDE benchmarks, the boosted PINN matches or exceeds the accuracy of standard monolithic PINNs while using substantially fewer parameters and, in stiff regimes, converges where standard PINNs fail outright.

As mentioned in the Discussion (Section \ref{sec:discussion}), these gains come with limitations. Convergence degrades beyond a certain stiffness threshold. Performance on long temporal horizons remains an open challenge shared with standard PINNs, and our second-order optimizers have so far only been validated on nonstiff ODEs. Extending second-order optimization to stiff equations and PDEs and addressing long-horizon training are natural directions for future work.

Taken together, the results, ablation studies, and sensitivity analysis support boosting as a principled and practical alternative to monolithic PINN training. This approach trades a single large network for a sequence of small, theoretically grounded corrections, with the potential to scale to harder physics-informed learning problems where standard architectures struggle.
\appendix

\section{Parameter Regimes: Duffing Equation}

Table~\ref{tab:duffing_param_sets} lists the ten parameter sets used to evaluate the Duffing equation. For each set, the Description column provides a qualitative characterization of the resulting dynamics. Results in the main text are averaged over all ten sets.

\begin{table}[ht]
    \centering
    \begin{tabular}{cccccccc l}
        \toprule
        ID & $\delta$ & $\alpha$ & $\beta$ & $\gamma$ & $\omega$ & $u_0$ & $\dot{u}_0$ & Description \\
        \midrule
        0 & 1.0 & -2.0 & 2.0  & 1.0 & 1.0 & 0.9 & 0.0 & double-well, moderate damping \\
        1 & 0.3 & -1.0 & 1.0  & 0.5 & 1.2 & 0.5 & 0.0 & classic chaotic regime \\
        2 & 0.1 &  1.0 & 1.0  & 0.3 & 0.8 & 0.6 & 0.0 & hardening spring, light damping \\
        3 & 2.0 & -1.0 & 1.0  & 0.5 & 1.0 & 1.0 & 0.0 & strongly damped, double-well \\
        4 & 0.5 &  1.0 & 0.05 & 0.5 & 1.0 & 0.5 & 0.0 & near-linear, moderate forcing \\
        5 & 0.4 &  1.0 & 1.0  & 0.8 & 2.5 & 0.3 & 0.0 & high-frequency forcing \\
        6 & 0.3 &  1.0 & 0.5  & 0.5 & 0.3 & 1.5 & 0.0 & low-frequency, large amplitude \\
        7 & 0.2 & -1.0 & 1.0  & 1.0 & 1.4 & 0.2 & 0.0 & double-well, strong forcing, near-chaotic \\
        8 & 0.5 &  1.0 & 4.0  & 0.8 & 1.2 & 0.7 & 0.0 & large nonlinearity, moderate damping \\
        9 & 0.0 &  1.0 & 1.0  & 0.0 & 1.0 & 0.4 & 0.0 & undamped, pure nonlinear oscillator \\
        \bottomrule
    \end{tabular}
    \caption{Duffing equation parameter regimes.}
    \label{tab:duffing_param_sets}
\end{table}

\section{Model Architecture}

All models take in the independent variables as input. The only exception is the Allen--Cahn (Periodic) network, which takes three inputs: $\cos(\pi(x+1))$ and $\sin(\pi(x+1))$, as shown in equation \ref{eq:phase_angle}. Replacing the raw spatial coordinate $x$ to encode the periodic domain $[-1,1]$ as a circle, guaranteeing that the network output satisfies the periodic boundary conditions $u(-1,t) = u(1,t)$ and $\partial_x u|_{x=-1} = \partial_x u|_{x=1}$. The third input is time $t$. 

\begin{table}[H]
    \centering
    \begin{threeparttable}

    \resizebox{\textwidth}{!}{%
    \begin{tabular}{llllll}
        \toprule
        \textbf{Boosted PINN} & \textbf{Neural Architecture} 
        & \textbf{WL Parameters} & \textbf{Activation} & \textbf{LR} \\
        \midrule
        Duffing Equation$^{\dagger}$  
            & $(t) - 32 - 32 - (u)$                                    
            & $2{,}209$  & $\sin$  & $1.0 \times 10^{-2}$ \\
        NRD$^{\dagger}$               
            & $(t) - 16 - 16 - (u)$                                    
            & $593$      & $\tanh$ & $1.0 \times 10^{-2}$ \\
        Lotka--Volterra               
            & $(t) - 32 - 32 - 32 - 32 - (x, y)$                      
            & $3{,}298$  & $\tanh$ 
            & $1.0 \times 10^{-3} \to 1.0 \times 10^{-5}$ \\
        Van der Pol                   
            & $(t) - 128 - 128 - 128 - 128 - (u)$                     
            & $49{,}921$ & $\sin$  
            & $1.0 \times 10^{-2} \to 1.0 \times 10^{-4}$ \\
        Burgers'                      
            & $(x, t) - 32 - 32 - 32 - (u)$                           
            & $2{,}241$  & $\sin$  
            & $1.0 \times 10^{-2} \to 1.0 \times 10^{-3}$ \\
        Allen-Cahn (Dirichlet)        
            & $(x, t) - 64 - 64 - (u)$                                
            & $4{,}417$  & $\tanh$ & $1.0 \times 10^{-2}$ \\
        Allen-Cahn (Periodic)        
            & $(\cos(\pi(x+1)),\, \sin(\pi(x+1)),\, t) - 64 - 64 - 64 - (u)$   
            & $8{,}641$  & $\tanh$ 
            & $1.0 \times 10^{-2} \to 5.0 \times 10^{-3}$ \\
        \bottomrule
    \end{tabular}}
    \caption{Boosted PINN weak learner architectures for different 
    equations.}
    \label{tab:architecture_boosted}
    \begin{tablenotes}
        \scriptsize
        \item[$\dagger$] No learning rate scheduler; learning rate is fixed throughout training.
    \end{tablenotes}
    \end{threeparttable}
\end{table}

\begin{table}[H]
    \centering
    \begin{threeparttable}
    \resizebox{\textwidth}{!}{%
    \begin{tabular}{lllll}
        \hline
        \textbf{Boosted PINN} & \textbf{Stages} & \textbf{Stage Weights} & \textbf{Epochs per Stage} & \textbf{Early Stopping} \\
        \hline
        Duffing Equation       & 20 & 0.01--0.055  & $180 \times 20^{\dagger}$                                    & False \\
        NRD                    & 20 & 0.05         & $70 \times 20^{\dagger}$                                     & False \\
        Lotka-Volterra         & 10 & 0.05         & $15000 \times 1$, $2000 \times 9$                            & True  \\
        Van der Pol            & 40 & 0.05         & $2000 \times 10$, $4000 \times 10$, $8000 \times 20$         & True  \\
        Burgers'               & 10 & 0.05--0.0275 & $2000 \times 5$, $4000 \times 5$                             & True  \\
        Allen-Cahn (Dirichlet) & 3  & 0.05         & 3000, 2500, 500                                              & True  \\
        Allen-Cahn (Periodic)  & 10 & 0.05         & $8000 \times 1$, $4000 \times 9$                             & True  \\
        \hline
    \end{tabular}}
    \caption{Boosted PINN training hyperparameters for different equations.}
    \label{tab:hyperparams_boosted}
    \begin{tablenotes}
        \scriptsize
        \item[$\dagger$] The notation $n \times k$ denotes $k$ consecutive stages each trained for $n$ epochs.
    \end{tablenotes}
    \end{threeparttable}
\end{table}

The total epoch budget for the standard PINN is equal to the sum of epochs across all boosting stages. For example, in the Allen--Cahn (Periodic) example, the standard PINN is trained for $8{,}000 + 4 \times 9{,}000 = 44{,}000$ epochs. This ensures the standard PINN is given at least as much training time as the full boosted ensemble.

We match the architecture of the standard PINN to that of the weak learners, rather than dividing the standard PINN's capacity across boosting stages. Fixing the base network architecture in this way isolates the training procedure as the sole variable of interest: given an identical network, the comparison asks whether it is more effective to optimize it jointly in a single training run or to train it sequentially as a series of residual correctors. Because the boosted model applies this same base architecture $K$ times, its total parameter count exceeds that of the standard PINN by a factor of $K$. We do not correct for this, as doing so would risk under-parameterizing each weak learner relative to the residual it must fit, introducing a confound of its own. Instead, we rely on time to convergence as the primary comparison metric, since it directly captures the added computational cost of training multiple weak learners sequentially. Any additional time required to train and aggregate the $K$ stages is reflected in the reported wall-clock time, ensuring the comparison remains fair with respect to actual computational cost rather than parameter count.

\begin{table}[H]
\centering
\begin{tabular}{lll}
    \toprule
    \textbf{Standard PINN} & \textbf{LR} & \textbf{Early Stopping} \\
    \midrule
    Duffing Equation        & $1.00 \times 10^{-3}$                          & Yes \\
    NRD                     & $5.00 \times 10^{-4}$                          & Yes \\
    Lotka--Volterra         & $1.00 \times 10^{-3}$                          & Yes \\
    Van der Pol             & $1.00 \times 10^{-3}$                          & Yes \\
    Burgers'                & $1.00 \times 10^{-2}$                          & Yes \\
    Allen-Cahn (Dirichlet)  & $1.00 \times 10^{-2}$                          & Yes \\
    Allen-Cahn (Periodic)   & $1.00 \times 10^{-2} \to 5.00 \times 10^{-3}$  & Yes \\
    \bottomrule
\end{tabular}
\caption{Standard PINN training hyperparameters.}
\label{tab:architecture_std}
\end{table}

\section{Sampling Strategy}
\label{section:sampling_strategy}
\begin{table}[H]
\centering
\begin{tabular}{lll}
    \toprule
    \textbf{Problem} & \textbf{Strategy} & \textbf{Validation Set Size ($N$)} \\
    \midrule
    Duffing Equation        & Sample from a fixed $N$ training set     & $3{,}000$   \\
    NRD                     & Sample from a fixed $N$ training set     & $6{,}000$   \\
    Lotka--Volterra         & Sample from a fixed $N$ training set     & $3{,}000$   \\
    Van der Pol             & Sample from a fixed $N$ training set     & $9{,}000$   \\
    Burgers'                & Sample from a fixed $N$ training set     & $1{,}000$   \\
    Allen-Cahn (Dirichlet)  & Sample a fresh $N$ set for each stage    & $800{,}000$ \\
    Allen-Cahn (Periodic)   & Sample a fresh $N$ set for each stage    & $51{,}456$  \\
    \bottomrule
\end{tabular}
\caption{Validation set size and sampling strategy for each problem.}
\label{tab:validation_set}
\end{table}

For most problems, the validation set is sampled once from a fixed set of $N$ points and held constant throughout training. For the Allen--Cahn examples, however, a fresh validation set of size $N$ is resampled at each boosting stage, consistent with the collocation-point resampling strategy described earlier for these two examples.

\section{Error and Residual Metrics}
\label{appendix:error_residual_metrics}

Let $u_{\text{pred}}$ denote the model prediction and $u_{\text{true}}$ the reference (analytical or high-fidelity numerical) solution, both evaluated on a discrete set of points $\{x_i\}_{i=1}^N$ (and possibly times $\{t_i\}_{i=1}^N$). Let $\mathcal{N}[u](x,t) = 0$ denote the governing differential equation.

\subsection{Mean Squared Error}
Following standard practice in the PINN literature, we report this quantity as the mean squared error (MSE), though it is technically computed as the root mean squared error (RMSE):
\begin{equation}
    \text{MSE} = \left(\frac{1}{N}\sum_{i=1}^{N} 
    |u_{\text{pred}}(x_i, t_i) - u_{\text{true}}(x_i, t_i)|^2
    \right)^{1/2}.
\end{equation}

\subsection{Residual Norm}
The residual norm measures the violation of the governing equation 
at the collocation points:
\begin{equation}
    \|\mathcal{R}[u_{\text{pred}}]\|_2
    = \left(\frac{1}{N}\sum_{i=1}^{N} 
    \big|\mathcal{R}[u_{\text{pred}}](x_i, t_i)\big|^{2}
    \right)^{1/2},
\end{equation}
where $\mathcal{R}[u_{\text{pred}}](x_i, t_i)$ denotes the differential operator (e.g., ODE or PDE residual) applied to the predicted solution at the collocation points. A smaller residual norm indicates that the predicted solution more closely satisfies the governing equation.

\subsection{Relative $L^2$ Error}

\begin{equation}
\text{relative } L^2 \text{ error} = \frac{\left(\sum_{i=1}^N |u_{\text{pred}}(x_i,t_i) - u_{\text{true}}(x_i,t_i)|^2\right)^{1/2}}{\left(\sum_{i=1}^N |u_{\text{true}}(x_i,t_i)|^2\right)^{1/2}}.
\end{equation}

\section{Efficiency Tables}
Table~\ref{tab:ode_efficiency} illustrates the computational efficiency of the boosted PINN versus the standard PINN across ODE benchmarks. The wall clock time refers to the total training time. The number of iterations refers to the number of epochs required to achieve MSE $< 10^{-2}$. Speed-up is computed relative to the standard PINN (Adam):
$$
\text{Speed-up (Iterations)} = \frac{\text{Standard PINN iterations}}{\text{Boosted PINN iterations}},
$$
$$
\text{Speed-up (Wall Clock)} = \frac{\text{Standard PINN time}}{\text{Boosted PINN time}}.
$$
For example, a value of $1.25$ in iterations indicates the standard PINN required $25\%$ more epochs than the boosted PINN. A value of $0.50$ in wall clock time indicates the standard PINN uses only 50\% of the boosted PINN training time. In either case, large values are good for the boosted PINN.

For the Duffing equation, the boosted PINN requires fewer iterations to converge, but has a longer wall clock time due to the overhead of computing the Hessian and other computational overhead. This includes computing the previous solution, and the full aggregated solution after each stage completes; deep-copying network states to enable rollback in the event of divergence during optimization; and warm-starting each new weak learner from the previous stage's converged weights. For all other examples, the standard PINN fails to converge under the same epoch budget. 

For the PDE examples, there was only one example for which the standard PINN converged. For this example, the boosted PINN achieves a faster convergence with equal iterations. 

\begin{table}[H]
    \centering
    \begin{threeparttable}
    \resizebox{\textwidth}{!}{%
    \begin{tabular}{llcccccc}
        \toprule
        & & \multicolumn{2}{c}{Boosted PINN} 
          & \multicolumn{2}{c}{Standard PINN} 
          & \multicolumn{2}{c}{Speed-up} \\
        \cmidrule(lr){3-4} \cmidrule(lr){5-6} \cmidrule(lr){7-8}
        Problem & Method 
            & Iterations & Wall Clock (s) 
            & Iterations & Wall Clock (s) 
            & Iterations & Wall Clock \\
        \midrule
        \multirow{3}{*}{Duffing}
            & Adam          & $702$        & $8.23$       & $1{,}530$  
                            & $3.25$       & $2.18\times$ & $0.40\times$ \\
            & Adam + CG     & $738$        & $7.03$       & $-$        
                            & $-$          & $2.07\times$ & $0.46\times$ \\
            & Adam + Newton & $407$        & $187.72$     & $-$        
                            & $-$          & $3.76\times$ & $0.02\times$ \\
        \midrule
        NRD (stiff)
            & Adam          & $858$        & $0.27$       
                            & $\dagger$    & $\dagger$    
                            & $\infty$     & $\infty$     \\
        \midrule
        \multirow{3}{*}{NRD (nonstiff)}
            & Adam          & $658$        & $0.23$       & $\dagger$  
                            & $\dagger$    & $\infty$     & $\infty$     \\
            & Adam + CG     & $434$        & $1.61$       & $-$        
                            & $-$          & $\infty$     & $\infty$     \\
            & Adam + Newton & $462$        & $40.28$      & $-$        
                            & $-$          & $\infty$     & $\infty$     \\
        \midrule
        Van der Pol (stiff)
            & Adam          & $134{,}234$  & $15{,}550.89$
                            & $\dagger$    & $\dagger$    
                            & $\infty$     & $\infty$     \\
        \midrule
        Lotka--Volterra
            & Adam          & $26{,}400$   & $38.06$      & $\dagger$  
                            & $\dagger$    & $\infty$     & $\infty$     \\
        \bottomrule
    \end{tabular}
    }
    \caption{Computational efficiency comparison between boosted PINN and standard PINN on ODE problems.}
    \label{tab:ode_efficiency}
    \begin{tablenotes}
        \scriptsize
        \item[$\dagger$] Standard PINN did not converge.
        \item[$-$] Experiment not conducted.
    \end{tablenotes}
    \end{threeparttable}
\end{table}

\begin{table}[H]
\begin{threeparttable}
    \centering
    \resizebox{\textwidth}{!}{%
    \begin{tabular}{llcccccc}
        \toprule
        & & \multicolumn{2}{c}{Boosted PINN}
          & \multicolumn{2}{c}{Standard PINN}
          & \multicolumn{2}{c}{Speed-up} \\
        \cmidrule(lr){3-4} \cmidrule(lr){5-6} \cmidrule(lr){7-8}
        Problem & Method
            & Iterations & Wall Clock (s)
            & Iterations & Wall Clock (s)
            & Iterations & Wall Clock \\
        \midrule
        Allen--Cahn ($D = 10^{-4}$, stiff)
            & Adam
            & $16{,}933$ & $499.44$
            & $\dagger$  & $\dagger$
            & $\infty$   & $\infty$ \\
        Allen--Cahn ($D = 1.0$, nonstiff)
            & Adam
            & $500$      & $2.78$
            & $500$      & $3.51$
            & $1.00\times$ & $1.26\times$ \\
        Burgers' ($\nu = 0.1$, nonstiff)
            & Adam
            & $600$      & $22.03$
            & $\dagger$  & $\dagger$
            & $\infty$   & $\infty$ \\
        Burgers' ($\nu = 0.002$, stiff)
            & Adam / L-BFGS
            & $-$        & $-$
            & $-$        & $-$
            & $-$        & $-$ \\
        \bottomrule
    \end{tabular}
    }
    \caption{Computational efficiency comparison between boosted PINN and standard PINN on PDE problems.}
    \label{tab:pde_efficiency}
    \begin{tablenotes}
        \scriptsize
        \item[$\dagger$] Standard PINN did not converge.
        \item[$-$] No converged solution obtained.
    \end{tablenotes}
\end{threeparttable}
\end{table}

\section{Coercivity and Monotonicity Implications}
\label{appendix:conditions}

\subsection{Existence of a Minimizer}
\label{appendix:conditions_existence}
Without any conditions on $F$, the infimum $$\inf_{u \in H^m(\Omega)} \|F(u)\|^2_{L^2(\Omega)}$$ might not be attained. A minimizing sequence $\{u_n\}$ could escape to infinity ($\|u_n\|_{H^m} \to \infty$) or converge weakly but not strongly.

Coercivity prevents this by ensuring that large residuals imply large norms:
$$
\|F(u)\|_{L^2} \to \infty \quad \text{as} \quad \|u\|_{H^m} \to \infty,
$$ 
so that minimizing sequences remain bounded in $H^m(\Omega)$. Existence then follows from the direct method of the calculus of variations \citep{evans-2010}:
\begin{itemize}
    \item Take a minimizing sequence $\{u_n\}$.
    \item Coercivity gives boundedness: $\|u_n\|_{H^m} \leq M$ for 
    some finite $M$.
    \item Reflexivity of $H^m(\Omega)$ gives a weakly convergent 
    subsequence $u_{n_k} \rightharpoonup u^*$.
    \item Weak lower semicontinuity of $\mathcal{L}$ gives 
    $\mathcal{L}(u^*) \leq \liminf_{k} \mathcal{L}(u_{n_k})$, so 
    $u^*$ is a minimizer.
\end{itemize}

\subsection{Uniqueness of the Minimizer}
\label{appendix:conditions_uniqueness}
Strong monotonicity is a standard condition in the theory of monotone operators (see \cite{bauschke-2017}). Without strong monotonicity, $F(u) = 0$ could have multiple solutions, and $\mathcal{L}$ could have multiple global minima. Recall from~\eqref{equation:monotonicity} that $F$ is assumed strongly monotone,
$$
    \langle F(u) - F(v),\, u - v \rangle_{L^2(\Omega)} 
    \geq \gamma \|u - v\|^2_{H^m(\Omega)}, 
    \quad \gamma > 0.
$$
This rules out multiple solutions. Suppose $F(u^*) = 0$ and $F(v^*) = 0$. Then:
$$
    0 = \langle F(u^*) - F(v^*),\, u^* - v^* 
    \rangle_{L^2(\Omega)} 
    \geq \gamma \|u^* - v^*\|^2_{H^m(\Omega)},
$$
which forces $\|u^* - v^*\|_{H^m(\Omega)} = 0$, and therefore $u^* = v^*$. Hence, the solution to $F(u) = 0$ is unique in $H^m(\Omega)$, and $\mathcal{L}$ has at most one global minimizer.

\bibliographystyle{plainnat}  
\bibliography{references}
\end{document}